\documentclass[letterpaper]{article} 
\usepackage{aaai23}  
\usepackage{times}  
\usepackage{helvet}  
\usepackage{courier}  
\usepackage[hyphens]{url}  
\usepackage{graphicx} 
\def\UrlFont{\rm}  
\usepackage{natbib}  
\usepackage{caption} 
\usepackage{algorithm}
\usepackage{algorithmic}

\usepackage{newfloat}
\usepackage{listings}
\DeclareCaptionStyle{ruled}{labelfont=normalfont,labelsep=colon,strut=off} 
\floatstyle{ruled}
\newfloat{listing}{tb}{lst}{}
\floatname{listing}{Listing}
\usepackage{algorithm}
\usepackage{algorithmic}

\usepackage{newfloat}
\usepackage{listings}
\usepackage{comment}
\usepackage{mathtools}
\usepackage{subcaption}
\usepackage{multirow}

\usepackage{xurl}
\usepackage{float}
\usepackage[shortlabels]{enumitem}
\setlist[enumerate]{nosep}
\setlist{nolistsep,leftmargin=2.0mm}
\usepackage{amsfonts}
\usepackage{xpatch}

\newcommand{\name}{{\sc{approxEDA }}}
\newcommand{\blink}{{\sc{BlinkDB }}}
\newcommand{\greedy}{{\sc{ciGreedy }}}

\iftrue

\newcommand{\tong}[1]{{\color{blue}{\small{\bf [Tong: #1]}}}}

\newcommand{\sm}[1]{{\color{brown}{\small{\bf [Subrata: #1]}}}}

\newcommand{\sg}[1]{{\color{purple}{\small{\bf [Shaddy: #1]}}}}

\else

\newcommand{\tong}[1]{}

\newcommand{\sm}[1]{}

\newcommand{\sg}[1]{}

\fi

\floatstyle{ruled}
\newfloat{listing}{tb}{lst}{}
\floatname{listing}{Listing}

\title{Reinforced Approximate Exploratory Data Analysis}
\author {
    Shaddy Garg\textsuperscript{\rm 1},
    Subrata Mitra\textsuperscript{\rm 1}\thanks{Corresponding Author},
    Tong Yu\textsuperscript{\rm 1},\\
    Yash Gadhia\textsuperscript{\rm 2}\thanks{Work done while at Adobe Research},
    Arjun Kashettiwar \textsuperscript{\rm 2†}
}
\affiliations {
    \textsuperscript{\rm 1} Adobe Research\\
    \textsuperscript{\rm 2} Indian Institute of Technology, Bombay\\
}

\begin{document}

\maketitle

\begin{abstract}

Exploratory data analytics (EDA) is a sequential decision making process where analysts choose subsequent queries that might lead to some interesting insights based on the previous queries and corresponding results.
Data processing systems often execute the queries on samples to produce results with low latency. 
Different downsampling strategy preserves different statistics of the data and have different magnitude of latency reductions. The optimum choice of sampling strategy often depends on the particular context of the analysis flow and the hidden intent of the analyst. 
In this paper, we are the first to consider the impact of sampling in interactive data exploration settings as they introduce approximation errors. We propose a Deep Reinforcement Learning (DRL) based framework which can optimize the sample selection in order to keep the analysis and insight generation flow intact. 
Evaluations with 3 real datasets show that our technique can preserve the original insight generation flow while improving the interaction latency, compared to baseline methods.

\end{abstract}
\section{Introduction}

Exploratory data analysis (EDA) is an interactive and sequential process of data understanding and insight generation where a user (e.g. a data analyst) issues an analysis action (i.e. a query) against a tabular data, receives some answers, subsection of the tabular data or some visualization and then decides which queries to issue next in order to understand the hidden characteristics of the data and associated insights better~\cite{ma2021metainsight, atena2020sigmod, edasim2018kdd}.
Characteristics and insights obtained using EDA are crucial for subsequent decision making in various domains~\cite{ma2021metainsight}.
%
%
The \textit{success} of an EDA session and the quality of insights obtained from it large depend on two things: (1) how interactive the system is (i.e. how quickly results are obtained) and (2) whether the outcome of previous sequence of queries are reliable and representative enough for an expert analyst to issue subsequent queries so as to uncover interesting characteristics. 
%
%
\begin{figure}[t]
\begin{center}
\includegraphics[width=0.85\columnwidth, height=0.25\columnwidth]{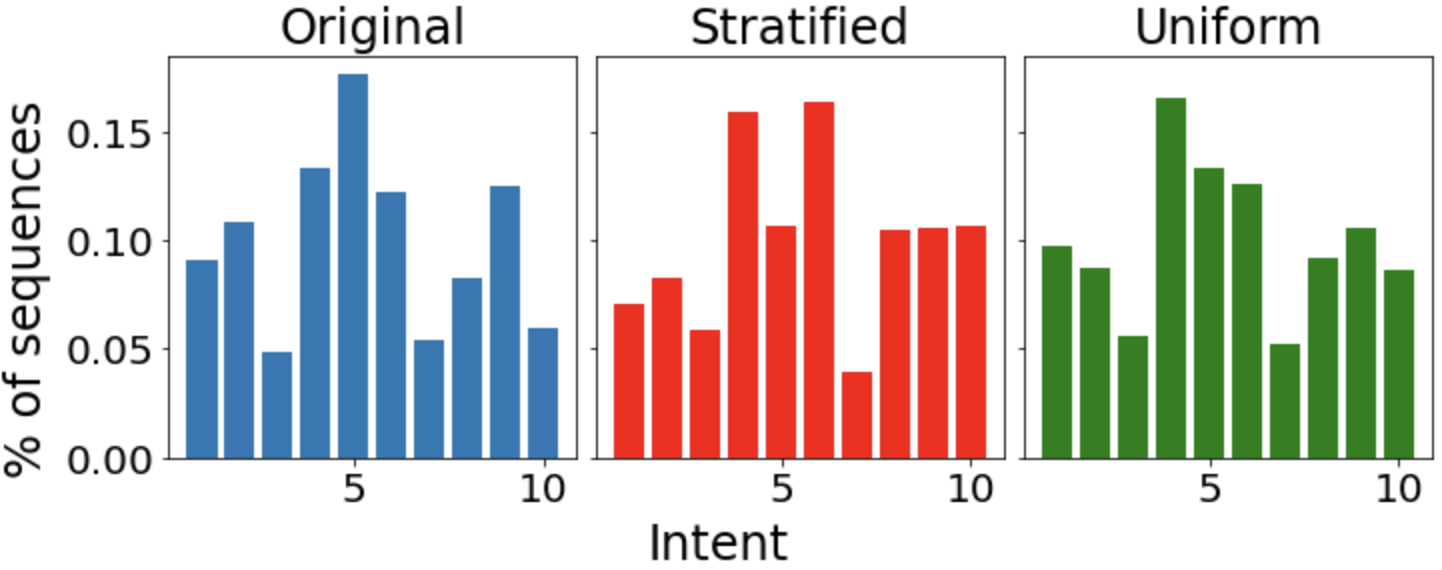}
     \caption{Divergence of intent distribution: Single sampling strategy (stratified or uniform) for all the contexts in sequential data exploration can make user to get mislead into wrong analysis flow due to approximation errors.} 
     \label{fig:intent_skew}
\end{center}
\end{figure}

To improve latency of interactions, downsampling the data has been proposed in various works target ting approximate query processing (AQP)~\cite{chaudhuri2017approximate, park2018verdictdb, sheoran2022aaai} and visualizations~\cite{park2016visualization, moritz2017trust, porwal2022sigmod}. For large datasets that are often a target for EDA~\cite{galakatos2017revisiting, wang2014sample}, running queries on samples can substantially reduce the query latency.
However, sampling creates approximation errors and can mislead the users in an interactive data exploration flow, especially when the use, choice and degree of approximation is transparent to the user.
In this paper we are the first to explore a setting where an EDA system \textit{transparently} uses different forms of downsampling strategy on the original data to run the queries but protect the user from being mislead due to approximation error from poor choice of samples. 
Figure~\ref{fig:intent_skew} shows distributions of 10 user-intents extracted from several thousand of realistic EDA sequences on Flights data.
It can be observed that if only a particular downsample strategy (e.g. stratified or uniform sampling) is used for every context in the entire sequence, the resulting intent distribution diverges from the ideal. 
%
%
This is because the results of the previous query can get distorted due to sampling and may prompt the user to a false path of analysis, which would have been useless if the original data was used. Second, there are numerous sampling techniques available, such as uniform sampling with different sampling rates, stratified sampling of few different kinds, systematic sampling, cluster sampling, diversity sampling etc.
Each of these sampling algorithms are good at preserving some properties of the data at the cost of introducing approximation errors for other properties. 
For a sequential and interactive data exploration workflow where multiple different types of queries are used on different subsections of the data, there is often not a single sampling strategy that wins over everything else. This is because, in a sequential interactive exploration the amount of approximation in the results or visualizations generated from the previous queries, influences the user decision on what next query to run. 
Moreover, there is a trade-off between query latency and accuracy, when sampling is used.
Choosing a sample that gives higher approximation error (while making the query faster) can be desirable, as long as the approximation error does not mislead the user to make wrong decisions about subsequent analyses flow or alter the key takeaways.  

This brings us to the second major problem that we handle in this paper. EDA being fundamentally an open-ended and expertise-driven process to understand a new dataset, there is no explicit \emph{intent} that can be attributed to the analysis flow.
Some recent papers looked at how to learn from EDA sequences performed by expert users in order to mimic those in a simulator to generate a sequence of queries and results on a new dataset, to help novice analysts~\cite{atena2020sigmod}.
Similarly~\cite{edasim2018kdd} provides suggestions to novice users about the next query to run based on learning from expert user sessions. 
While these works identify that exploratory analysis is not exactly random in nature and expert analysts usually follow few core implicit intents, but they do not handle how intent-aware learning can be performed in such situations. 
This lack of clear intent, as well as not having a clear measure of success or a implicit/explicit feedback at the end also makes our problem markedly distinct from other interactive settings such as task-oriented dialog systems \cite{liu2018dialogue,shi2019build}, conversational recommender systems \cite{lei2020interactive,zhang2020conversational} and interactive retrieval systems \cite{guo2018dialog}. When developing the learning agent for these systems, it is usually assumed that the users have a clear intent, from which the loss function or reward function can be naturally derived. However, in our task the users usually have implicit intent, which makes development of the reward function non-trivial. 
%

To jointly address the above two challenges, we propose an intent-aware deep reinforcement learning (DRL) framework, \name, to enable the interactive exploratory data analysis in the presence of approximation errors. 
%
In this paper, we focus on how to prevent intent-divergence (\emph{i.e.}, approximation errors misleads users to a different intent). It is assumed that there is limited \textit{intent-shift} (\emph{i.e.}, change of user's core intent characteristics) in the problem setting. 
This assumption is widely adopted in other interactive applications \cite{liu2018dialogue,guo2018dialog,tan2019drill}, although it is interesting to address the intent-shift \cite{xie2021tiage,arnold2019sector}. As the first work on approximate exploratory data analysis in the interactive setting, we leave addressing \textit{intent-shift} as future work.

Our contributions are summarized as follows.
\begin{itemize}
    \item To our best knowledge, we are the first to identify an important yet practical problem where approximations should be used to improve interactivity in sequential exploration of data but can potentially mislead analysts to wrong analyses paths or intent-divergence. 
    \item We make a case that depending on the context of the sequential analysis, different sampling strategies would be optimal for providing lower interaction latency while protecting intent-divergence.   
    \item We novelly formulate the problem as a reinforcement learning task. Our formulation captures both the computational cost of executing the queries, as well as the sequential decisions made by users for different implicit intents. We model the choice of down-sampling strategy as the action-space and optimize the divergence of sequential interactions by an user along with the interactivity of the system, in the presence of approximations, in an intent-aware manner.
    To efficiently optimize the agent without explicit user intent, we develop a novel reward function and DRL learning stop criteria.
    \item We show the effectiveness of our technique by extensively evaluating our solution on 3 real-world datasets and against several baselines. 
    We also present detailed ablation study and make our code and data available~\cite{open_repo}.
    
\end{itemize}

\begin{figure*}[t]
\includegraphics[width=0.98\linewidth]{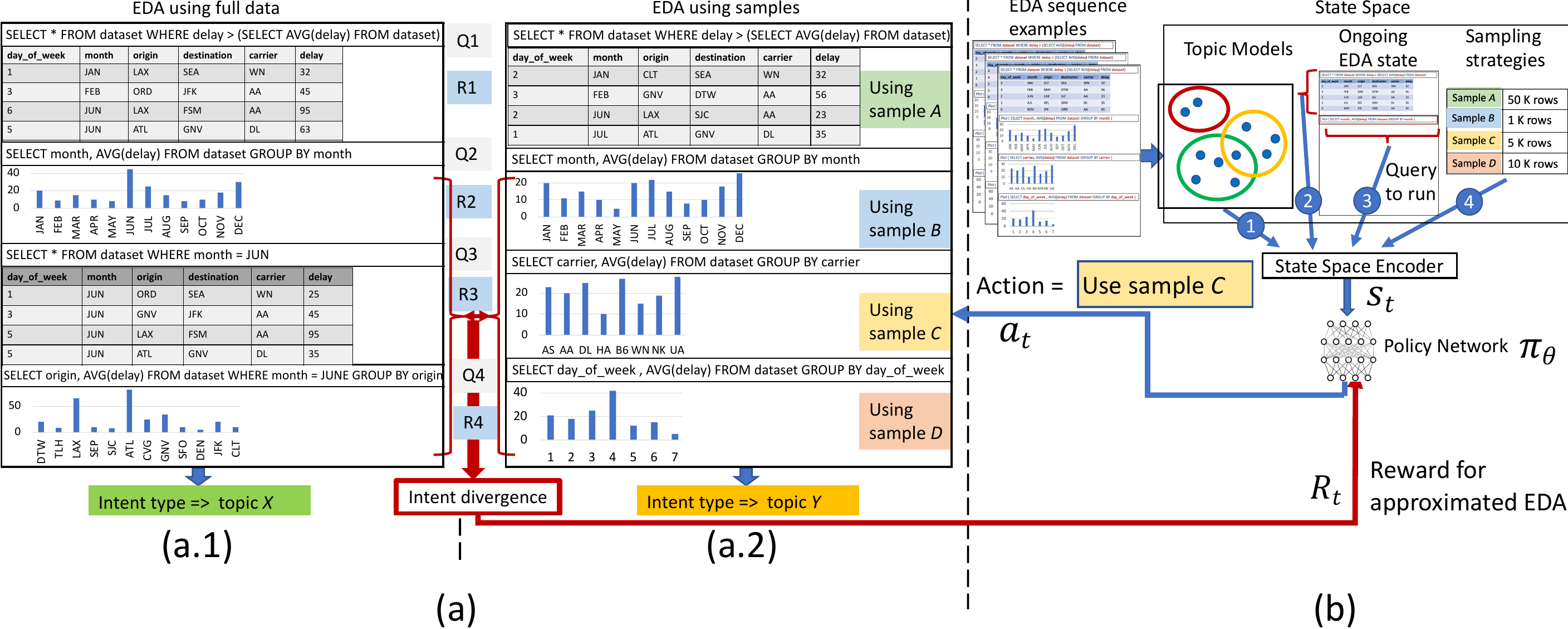}
     \caption{\textbf{(a)} Example of intent divergence between EDA sessions using full data \textbf{(a.1)} vs. using samples \textbf{(a.2)}. 
     Bad sample selection in (a.2) led to large errors in the result \textit{R2}. Consequently, user diverted from ideal query sequence and finally found misleading insight in \textit{R4}.
     \textbf{(b)} Workflow of our RL-based \name is shown. The agent transparently interacts with a live EDA session (e.g. one on the left) and chooses the best action $a_t$, i.e. the optimal sample to use for the current so that intent divergence is minimized. \name's \textit{State Space Encoder} combines 4 different information into a state $s_t$ that is used by the policy-network ($\pi_{\theta}$) to decide the best action. 
     }
     \label{fig:intent_divergence}
\vspace{-0.8em}
\end{figure*}

\section{Background and Motivations}

\subsection{Motivating Example}


In Figure~\ref{fig:intent_divergence}(a.1), we show example of an EDA being performed on \textit{all} the records of Flights data~\cite{Flights}. At first an analyst attempts to understand the flights delay and filters the entries that had delay higher than average (with query \texttt{Q1} and corresponding result/visualization \texttt{R1}). 
Then she wants to understand delays in detail by grouping those by \textit{month} (i.e. \texttt{Q2}) and observes from the bar-plot (\texttt{R2}) that month of \texttt{JUN} contributed highest delays.
Then she inspects by filtering the rows only for \texttt{JUN} (\texttt{Q3} and \texttt{R3}). Not finding anything obvious she groups these filtered results by \texttt{origin} airport (\texttt{Q4}) and from the result (\texttt{R4}) concludes that delays in flights originating from \texttt{LAX} and \texttt{ATL} airports is the main culprit. Then she might forward this \textit{insight} and investigate the root-cause of delays in these airports further. 
Please note: there is a lack of explicit intents~\cite{milo2020automating, ma2021metainsight} from the analysts corresponding to an EDA session.

Now in Figure~\ref{fig:intent_divergence}(a.2), we illustrate the problem of arbitrary sample selection if sampled are used to speed-up queries but with approximate results.
While nothing detrimental happens in the result \texttt{R1} and the analyst would still run (\texttt{Q2}) to understand the delays grouped by \texttt{month}, a poor choice of sample may not highlight the large amount of delay contributed by the month of
\texttt{JUN}. 
Therefore, not finding anything that stands out over different months, the analyst might go on to look for other anomalies or patterns of delay for different careers (\texttt{Q3} and \texttt{R3}) and further not finding anything that suspicious, attempts to understand delays over different \texttt{days\_of\_week}(\texttt{Q4} and \texttt{R4}) and concludes with a \textit{misleading insight} that delays are primarily caused of flights on \textit{$4^{th}$ day of the week}.
Here the analyst diverged from the ideal analysis flow (if original data was used) because of approximation errors in queries/results \texttt{Q1/R1} and \texttt{Q2/R2}. 
She finally made a wrong decisions on her choice of subsequent queries at \texttt{Q3}.   
As a result overall EDA flow diverged. 

Considering the above example, appropriate sub-sample selection is crucial based on the context of EDA at each step. Thus balancing low latency for good interactive experience vs. intent and insight preservation is the goal of this paper. 

\subsection{Sampling Strategies}

A sample dataset $T_s$ is a subset of rows from an original dataset $T$. Following different sampling strategies select these subsets differently with different set of parameters that control the size and statistics of $T_s$.
We consider sampling strategies listed in Table~\ref{tab:actions} with different associated parameters that control the sizes of the subsamples and consequently the amount of statistical information. More details of the sampling strategies is in Appendix \textbf{\ref{app:sampling_details}}.

\begin{table}
\scalebox{0.8}{
 \begin{tabular}[ht!]{|l|l|l|}
    \hline
    {Sample Name} & {Short Name} & {Parameters} \\
    \hline
    {Uniform} & {Uni@$\tau$} & {$\tau$= [0.01, 0.05, 0.1]} \\
    \hline
    {Systematic} & {Sys@$k$} & {$k$= [100, 20, 10]}  \\
    \hline
    {Proportional stratified} & {Strat@$\tau$} & {$\tau$= [0.01, 0.05, 0.1]} \\
    \hline
    {At most $K$ stratified} & {Strat-K@$K$} & {$K$= [2k, 10k, 20k]} \\
    \hline
    {Cluster} & {Clus@$k$} & {$k$= 10, $\tau$= [0.01, 0.05, 0.1]} \\
    \hline
    {MaxMin Diversity} & {MaxMin@$k$} & {$k$= 0.1*$|T|$} \\
    \hline
    {MaxSum Diversity} & {MaxSum@$k$} & {$k$= 0.1*$|T|$} \\
    \hline
    \end{tabular}
    }
 \captionof{table}{Sampling strategies and corresponding parameters that together creates our action space. We use the \textit{Short Name} to refer these.}
 \label{tab:actions}
\end{table}

\vspace{-0.1em}
\section{Overview of Proposed Approach}
In this paper we propose an intent-aware deep reinforcement learning (DRL)-based technique: \name{} that can make the optimal choice of samples to be used at each step of the analysis flow to speed interactions but prevents intent-divergence. 

\subsection{Design of \name{}}
We show the high-level system architecture of \name{} in Figure~\ref{fig:intent_divergence}(b).
\name{} first takes in a history of EDA sequences performed by human data analysts and uses Topic Modeling technique to identify a set of implicate intents in those sequences. 
Then the RL-agent of \name{} takes into account:
(1) these clusters of these latent intents,
(2) the history of the ongoing EDA exploration session including the query used so far and the corresponding display-outputs (graphs and dataframes),
(3) the next query the user is intending to run,
and (4) the set of available samples created with different sampling strategies along with the size of each. 
The RL-agent, which is parameterized by a deep neural network~\cite{lillicrap2015, mnih2015human}, is trained \textit{offline} to choose the optimal sampling strategy as the best \textit{action} for different context of the analyses and intent. We considered alternate RL algorithms (DQN [\cite{mnih2013playing}] and REINFORCE [\cite{sutton2018reinforcement}]) but found the convergence and stabilization of A2C training (Lillicrap et. al), which is a hybrid between value and policy based approach, was better and reduces variance.
The best action corresponding to each step, i.e. for each query in the EDA session attempts to minimize the divergence of intents due to approximation error caused by different samples, while optimizing for lower latency. 


For our problem, lack of explicit intents in the EDA sessions, a moderately large action space of possible sampling strategies combined with the choice of parameters for each of those strategies that would control the statistical information preserved in the samples and associated trade-off for latency based on the size of the samples ($\#$ of rows), makes this a complex optimization problem. 
Appendix\textbf{~\ref{app:discussion}} further discusses why RL better choice instead of greedy solutions.

\subsection{Intent Identification}
\label{sec:intent_identification}

The problem of intent aware sample selection requires us to first solve the problem of identifying the intent of the user given a query sequence $Q_j$. 
%
%
Similar to analyzing natural language texts to understand the topics, we can analyze query sequences to discover the user intents. Inspired by the previous work which uses topic modeling techniques to model goals
that users carry out by executing a sequence of commands \cite{aggarwal2020goal}, we leverage topic modeling to discover the user intent, which is further used to derive the reward signal as detailed later in Section \ref{sec:reward}. 


Specifically, we use a bi-term model (BTM) \cite{btm2013www} for the identification of the intent. Each topic output from the BTM model corresponds to an intent. For a query sequence $Q_j$, the output from the BTM model, \(\phi\) is the probability distribution over all the topics.

{\small
\begin{equation}
    \phi(Q_j) = \{P(t=t_i | Q=Q_j)\}_{i=1}^kK
\end{equation}
}
Here $K$ is the number of topics, and each $t_i$ corresponds to a topic. To get the corresponding intent, we take the $argmax$ over the probability distribution. The obtained intent, $I_{Q_j}$ is considered to be the user intent by \name{}.

{\small
\begin{equation}
    I_{Q_j} = argmax(\{P(t=t_i | Q=Q_j)\}_{i=1}^K)
\end{equation}
}
To decide the optimum number of topics for the BTM model, we use a coherence evaluation metric, the UCI measure~\cite{newman2010evaluating}, defined as

{\small
\begin{equation}
    score_{UCI}(q_i, q_j) = \log{\frac{p(q_i, q_j)}{p(q_i)p(q_j)}},
\end{equation}
}
where $p(q_i)$ represents the probability of seeing a query $q_i$ in a query sequence and $p(q_i, q_j)$ is the probability of observing both $q_i$ and $q_j$ co-occurring in a query sequence computed as follows,$p(q_i) = \frac{M(q_i)}{M} \text{ and } p(q_i, q_j) = \frac{M(q_i, q_j)}{M}$, where $M(q_i)$ is the count of query sequences containing the query $q_i$, $M(q_i, q_j)$ is the count of sequences containing both $q_i, q_j$ and $M$ is the total number of query sequences.
The $UCI$ score for an intent $I$, is calculated as, $mean\{score_{UCI}(q_i,q_j | q_i, q_j \in I, i \neq j)\}$ and the overall UCI score the average across all intents. Higher UCI score is indicative of better grouping because one would expect two queries belonging to the same intent to show up together regularly in the query sequences. We compute the overall UCI score for all $K \in [2,15]$ and choose the best K.



\section{RL Formulation}

We use RL for intent-aware sample selection the agent learns the optimal policy modeled as a conditional distribution $\pi(a|s)$, specifying the probability of choosing action $a \in A$ when in state $s \in S$. 
If the agent chooses an action $a \in A$ at state $s \in S$, it receives a reward $r(s,a)$.

\subsection{Action Space}
\vspace{-0.2em}
A set of $n$ samples are pre-created using different sampling strategies (and of different sizes). 
These samples are the set of $n$ actions: $A = \{a_1, a_2, .., a_i\}_{i=1}^n$ for the RL-agent.

\subsection{State Space}
As shown in Figure~\ref{fig:intent_divergence} (b), we use 4 components while deciding which action to choose. These components form the state space of the RL agent. The agent at each timestep $t$ for a query $q_t$ has access to the ongoing EDA state comprising of the previous $k$ queries $\{q_{t-1}, q_{t-2}, ..., q_{t-k}\}$ and their result vectors $\{v_{t-1}, v_{t-2}, ..., v_{t-k}\}$. 
Since reducing the latency is one of the primary goals of the agent, it also takes into account the latency encountered till timestep $t$. We denote this latency factor by $C_t=\sum_{i=0}^t c_i$, where $c_i$ is the latency associated with running a query $q_i$ on a sample $a_i$. Finally, to preserve the intent information of the sequence, we also include the probability distribution over all the topics, $I_t = \phi(Q_t)$ as one of the state space components. Formally, the state space can be written as, 
{\small
\begin{equation}
\label{eq:state_space}
    s_t = \{ ((q_{t}, v_{t}),(q_{t-1}, v_{t-1}), (q_{t-2},v_{t-2})), I_t, C_t  \}
\end{equation}
}

Once the agent chooses an action $a_t \in A$, the agent moves to the next state $s_{t+1}$ and returns a result (display vector) $v_t$, as the result of the query $q_t$. Our goal here is to choose $a_t$ at each step, such that the overall latency for the sequence is minimised, simultaneously preserving the intent of the generated sequence. To do this we develop several reward functions and propose an intent-aware RL framework based on A2C methodology~\cite{mnih2016asynchronous}.
Details of our training algorithm is given in supplement (\ref{app:training}).

\subsection{Reward Design}
\label{sec:reward}
\vspace{-0.2em}
\name{} calculates the reward at the end of each exploratory session. For the calculation of reward, we use a set of sequences generated on the original table and we refer them as ground truth, $O_i \in O$. Each $O_i$ is a sequence of queries. Similarly, the sequence generated by the agent at the end of each episode is represented as $Q_t$.
Reward design combines following components.

\noindent\textbf{Latency Reward. }
Choosing samples of smaller size gives lower latency and hence this reward is calculated as:

{\small
\begin{equation}
\label{latency_reward}
    R_{latency} = \sum_{t=1}^{l} r(s_t, a_t),
\end{equation}
}
,where $r(s_t, a_t) = 1 - \frac{|{T_s}|}{|T|}$, where $T_s$ is the time taken to run on a sample, $T$ is the time taken to run on the original dataset, and $l$ is the length of the sequence in the session. 
This reward helps to navigate the trade-off between lower approximation error vs. lower latency. 

\noindent\textbf{Intent Reward. }
Intent reward is designed to prevent divergence of EDA sequences due to approximations (\S!\ref{sec:problem}). 
To prevent intent-divergence we introduce two components in the intent reward. We denote the distance function for distance between two query sequences corresponding to two EDA sessions to be $Distance(Q_1, Q_2)$. We define the reward function to be,

{\small
\begin{equation}
\label{eq_eda_sim}
    R_{dis} = 1 - Distance(Q_t, Q_{ground}).
\end{equation}
}
where $Q_{ground}$ is the closest sequence to the given sequence $Q_{t}$ in the original set $O$.
 We use \texttt{EDA-Sim} defined in \cite{atena2020sigmod} as distance metric. EDA-Sim compares the content and order of the results from the query sequence to calculate a similarity score. Based on this score we find a sequence from $O$ that is closest with current sequence.

However, it is possible that two sequences may not have the exact same ordering or content (resulting in low EDA-Sim score) but still overall have same underlying intent. 
Hence we also use Euclidean Distance ($EuD$) to compare intent identified from topic model distributions (\S~\ref{sec:intent_identification}) as follows:

{\small
\begin{equation}
\label{topic_reward}
    R_{topic} = 1 - EuD(\phi(Q_t), \phi(Q_{ground}))
\end{equation}
}

Full intent reward is: $R_{intent} = R_{dis} + \delta*R_{topic}$.

\noindent\textbf{Termination Reward. }
While the whole EDA session, comprising of query and corresponding visualization sequences, are important to the analyst, usually last few queries are the most important as they provide key takeaway insights. 
%
%
Therefore, the another component of our reward is based on final insight preservation. The first subcomponent $R_{match} = \{0, 1\}$, a binary 
reward if the last $k$ queries of the generated sequence $Q_t$ match with at least one of the ground truth sequence belonging to the same intent cluster.
However, often different query sequence lead to similar insights. Therefore, the other subcomponent  $R_{recall}$ as the top-k recall for the display vectors (results) from the generated sequence, $Q_t$ and the closest sequence identified, $Q_{ground}$ in Eqn.~\ref{eq_eda_sim}.
Full termination reward is: $R_{term} = R_{match} + R_{recall}$.

%

Complete reward is $R_t = R_{latency} + \beta*R_{intent}+\gamma*R_{term}$.

\noindent\textbf{Intent-divergence vs. intent-shift}
This paper addresses the problem of \textit{intent-divergence} where the users' have a set of consistent implicit objectives/intents during the EDA. 
While \name{} deliberately introduces a bias towards a set of intents inferred from historical analysis sequences, this is justified as real expert users often follow some standard exploration strategies and look for similar types of insights even on new datasets as observed by multiple prior works~\cite{atena2020sigmod, jain2016sqlshare, milo2018next} and do not perform random explorations.
However, if such implicit objectives suddenly change causing an \textit{intent-shift}, then \name{}'s current design would not be able to detect that. 
In this paper, it is assumed that there is limited \textit{intent-shift} (\emph{i.e.}, change of user's core intent characteristics). 
This assumption is widely adopted in other interactive applications \cite{liu2018dialogue,guo2018dialog,tan2019drill}, although it is interesting to address the intent-shift \cite{xie2021tiage,arnold2019sector}. As the first work on approximate exploratory data analysis in the interactive setting, we leave addressing \textit{intent-shift} as future work.

\subsection{Implementation \& Deployment}

\begin{algorithm}[th]
        \begin{algorithmic}[1]
        \label{alg:alg1}
        \caption{\name{}'s Training Algorithm}
        \STATE Train the ATENA~\S{5.2} on given dataset $D$
        \STATE Generate sequences $O$ using ATENA
        \STATE Train a $BTM$ model $\phi$ with $K$ topics on $O$
        \STATE Create sample set $A=\{a_1,a_2,...,a_i\}_{i=1}^n$
        \STATE Initialise A2C model
        \FOR{$i = 1,...,episodes$}
            \STATE Initialize the state space $s_t$ as in (\ref{eq:state_space}) with zeroes
            \FOR{$t = 1,...,l$}                    
                    \STATE Generate a query $q_t$ from the simulator
                    \STATE Take action $a_t$ for the given $s_t$
                    \STATE Update $s_t$ with latest query $q_t$, display vectors $v_t$, latency $C_t$ and current intent as $\phi(Q_t)$
            \ENDFOR
            \STATE Calculate Individual rewards using~\S{4.3}.
            \STATE Calculate the total reward $R_t$
            \STATE Calculate $J_R(\pi) = \mathbb{E}_{a_t \sim \pi_\theta}[R_t]$
            \STATE Compute loss functions $\mathbb{L}(\theta)$ for policy network
            \STATE Compute $\mathbb{L}(\omega)$ for value network
            \STATE Update A2C model
        \ENDFOR
        \end{algorithmic}
    \end{algorithm}

Algorithm 1 explains the steps we use for training our RL Agent. It takes in the dataset $D$, processes it by creating samples as shown in line 4 and training ATENA simulator on the same as shown in line 1. Then on the generated sequences, a BTM model is trained in line 3. Once this is done, we then start the training of our agent as demonstrated in lines 5-18 of Algorithm 1. We follow episodic training where we calculate rewards at the end of each episode as shown in line 13. Once we calculate the total reward, we compute the loss functions and update both actor (policy) and critic (value) networks as explained in \S{\ref{apps:model_implementation}}. Appendix \textbf{\ref{app:discussion}} discusses deployment aspects.

\section{Experiments and Analysis}
Our experiments aim to answer these:

\begin{itemize}
    \item \textbf{RQ1.} Can \name prevent intent-divergence by context-aware sample selection? 
    (Results in \S~\ref{sec:eval_intent})
    \item \textbf{RQ2.} Can approximate EDA with \name still captures the final insights as the original?
    (Results in \S~\ref{sec:eval_insight})
    \item \textbf{RQ3.} Can \name provide significant latency reduction while addressing RQ1 and RQ2?
    (Results in \S~\ref{sec:eval_latency})
\end{itemize}

\begin{figure}[!t]
\centering
\begin{minipage}[t]{0.495\columnwidth}
\vspace{0pt}
\scalebox{0.65}{
\begin{tabular}{|l|l|l|}
\hline
\begin{tabular}[c]{@{}l@{}}Sampling \\ Strategy\end{tabular} & \begin{tabular}[c]{@{}l@{}}No. of \\ Rows\end{tabular} & \begin{tabular}[c]{@{}l@{}}Effective\\  SR\end{tabular} \\ \hline
Uni@1\%                                                      & 5232                                                   & 0.01                                                    \\ \hline
Uni@5\%                                                      & 26155                                                  & 0.05                                                    \\ \hline
Uni@10\%                                                     & 52312                                                  & 0.1                                                     \\ \hline
Sys@100                                                      & 5231                                                   & 0.01                                                    \\ \hline
Sys@20                                                       & 26154                                                  & 0.05                                                    \\ \hline
Sys@10                                                       & 52311                                                  & 0.1                                                     \\ \hline
Clus@1\%                                                     & 5230                                                   & 0.01                                                    \\ \hline
Clus@5\%                                                     & 26154                                                  & 0.05                                                    \\ \hline
Clus@10\%                                                    & 52306                                                  & 0.1                                                     \\ \hline
MinMax@5\%                                                   & 27202                                                  & 0.052                                                   \\ \hline
MaxSum@5\%                                                   & 25109                                                  & 0.048                                                   \\ \hline
KStrat1@2k                                                   & 14124                                                  & 0.027                                                   \\ \hline
KStrat1@10k                                                  & 28771                                                  & 0.055                                                   \\ \hline
KStrat1@20k                                                  & 57542                                                  & 0.11                                                    \\ \hline
KStrat2@2k                                                   & 9416                                                   & 0.018                                                   \\ \hline
KStrat2@10k                                                  & 25125                                                  & 0.048                                                   \\ \hline
KStrat2@20k                                                  & 47080                                                  & 0.09                                                    \\ \hline
Strat1@1\%                                                   & 5220                                                   & 0.01                                                    \\ \hline
Strat1@5\%                                                   & 26146                                                  & 0.05                                                    \\ \hline
Strat1@10\%                                                  & 52375                                                  & 0.1                                                     \\ \hline
Strat2@1\%                                                   & 5232                                                   & 0.01                                                    \\ \hline
Strat2@5\%                                                   & 26148                                                  & 0.05                                                    \\ \hline
Strat2@10\%                                                  & 52320                                                  & 0.1                                                     \\ \hline
Strat3@1\%                                                   & 5240                                                   & 0.01                                                    \\ \hline
Strat3@5\%                                                   & 26150                                                  & 0.05                                                    \\ \hline
Strat3@10\%                                                  & 52315                                                  & 0.1                                                     \\ \hline
Strat4@1\%                                                   & 5235                                                   & 0.01                                                    \\ \hline
Strat4@5\%                                                   & 26149                                                  & 0.05                                                    \\ \hline
Strat4@10\%                                                  & 52321                                                  & 0.1                                                     \\ \hline
\end{tabular}
}
\captionof{table}{Details of Samples generated on Flights Dataset}
\label{tab:flights_samples}
\end{minipage}%
\hfill
\begin{minipage}[t]{0.495\columnwidth}
\vspace{0pt}
\begin{minipage}[t]{1.0\columnwidth}
\begin{minipage}[t]{1.0\columnwidth}
\vspace{0pt}
\scalebox{0.65}{
\begin{tabular}{|l|l|l|}
\hline
\begin{tabular}[c]{@{}l@{}}Sampling \\ Strategy\end{tabular} & \begin{tabular}[c]{@{}l@{}}No. of \\ Rows\end{tabular} & \begin{tabular}[c]{@{}l@{}}Effective\\  SR\end{tabular} \\ \hline
Uni@1\%                                                      & 3242                                                   & 0.01                                                    \\ \hline
Uni@5\%                                                      & 16210                                                  & 0.05                                                    \\ \hline
Uni@10\%                                                     & 32419                                                  & 0.1                                                     \\ \hline
Sys@100                                                      & 3241                                                   & 0.01                                                    \\ \hline
Sys@20                                                       & 16209                                                  & 0.05                                                    \\ \hline
Sys@10                                                       & 32419                                                  & 0.1                                                     \\ \hline
Strat1@1\%                                                   & 3246                                                   & 0.01                                                    \\ \hline
Strat1@5\%                                                   & 16225                                                  & 0.05                                                    \\ \hline
Strat1@10\%                                                  & 32425                                                  & 0.1                                                     \\ \hline
Strat2@1\%                                                   & 3238                                                   & 0.01                                                    \\ \hline
Strat2@5\%                                                   & 16215                                                  & 0.05                                                    \\ \hline
Strat2@10\%                                                  & 32426                                                  & 0.1                                                     \\ \hline
Strat3@1\%                                                   & 3240                                                   & 0.01                                                    \\ \hline
Strat3@5\%                                                   & 16208                                                  & 0.05                                                    \\ \hline
Strat3@10\%                                                  & 32420                                                  & 0.1                                                     \\ \hline
Strat4@1\%                                                   & 3248                                                   & 0.01                                                    \\ \hline
Strat4@5\%                                                   & 16206                                                  & 0.05                                                    \\ \hline
Strat4@10\%                                                  & 32425                                                  & 0.1                                                     \\ \hline
\end{tabular}
}

\captionof{table}{Details of Samples generated on Housing Dataset}
\label{tab:housing_samples}
\end{minipage}
\begin{minipage}[t]{1.0\columnwidth}
\scalebox{0.65}{
\begin{tabular}{|l|l|l|}
\hline
\begin{tabular}[c]{@{}l@{}}Sampling \\ Strategy\end{tabular} & \begin{tabular}[c]{@{}l@{}}No. of \\ Rows\end{tabular} & \begin{tabular}[c]{@{}l@{}}Effective\\  SR\end{tabular} \\ \hline
Uni@1\%                                                      & 4266                                                   & 0.01                                                    \\ \hline
Uni@5\%                                                      & 21334                                                  & 0.05                                                    \\ \hline
Uni@10\%                                                     & 42670                                                 & 0.1                                                     \\ \hline
Sys@100                                                      & 4266                                                  & 0.01                                                    \\ \hline
Sys@20                                                       & 21334                                                & 0.05                                                    \\ \hline
Sys@10                                                       & 42669                                                  & 0.1                                                     \\ \hline
Strat1@1\%                                                   & 4267                                                   & 0.01                                                    \\ \hline
Strat1@5\%                                                   & 21336                                                  & 0.05                                                    \\ \hline
Strat1@10\%                                                  & 42672                                                  & 0.1                                                     \\ \hline
Strat2@1\%                                                   & 4235                                                   & 0.01                                                    \\ \hline
Strat2@5\%                                                   & 21330                                                  & 0.05                                                    \\ \hline
Strat2@10\%                                                  & 42664                                                  & 0.1                                                     \\ \hline
Strat3@1\%                                                   & 4235                                                   & 0.01                                                    \\ \hline
Strat3@5\%                                                   & 21340                                                  & 0.05                                                    \\ \hline
Strat3@10\%                                                  & 42667                                                  & 0.1                                                     \\ \hline
Strat4@1\%                                                   & 4237                                                   & 0.01                                                    \\ \hline
Strat4@5\%                                                   & 21367                                                  & 0.05                                                    \\ \hline
Strat4@10\%                                                  & 42673                                                  & 0.1                                                     \\ \hline
\end{tabular}
}

\captionof{table}{Details of Samples generated on Income Dataset}
\label{tab:income_samples}
\end{minipage}
\end{minipage}
\end{minipage}
\end{figure}

\subsection{Experimental Setup}
All experiments used a 32 core Intel(R)Xeon(R) CPU E5-2686 with 4 Tesla V100-SXM2 GPU(s). Model and implementation details are in supplement \ref{apps:model_implementation}.

\begin{figure*}[!t]
\begin{minipage}[t]{0.70\textwidth}
\centering
\begin{minipage}[t]{0.32\textwidth}
\begin{subfigure}{\textwidth}
\includegraphics[width=1\linewidth]{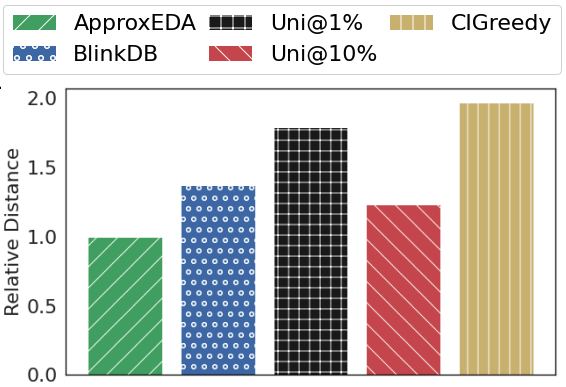}
\caption{Flights Dataset}
\end{subfigure}
\begin{subfigure}{\textwidth}
\includegraphics[width=1\linewidth]{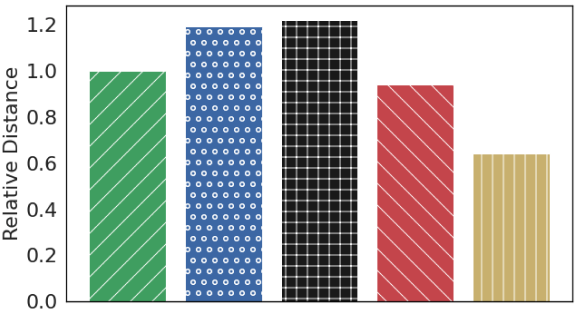}
\caption{Housing Dataset}
\end{subfigure}
\begin{subfigure}{\textwidth}
\includegraphics[width=1\linewidth]{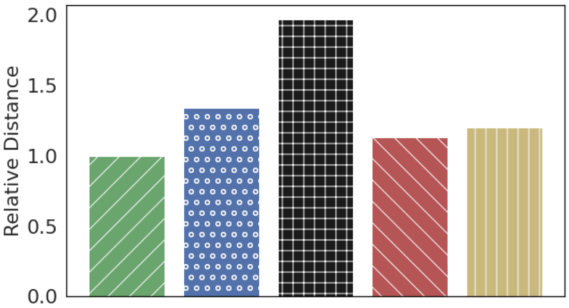}
\caption{Income Dataset}
\end{subfigure}
\caption{Intent Divergence: Euclidean distance between original intent distribution and methods using samples. Normalized w.r.t. \name. Lower is better.}
\label{fig:rel_dist_1}
\end{minipage}%
\hfill
\begin{minipage}[t]{0.32\textwidth}
\begin{subfigure}{1\textwidth}
\includegraphics[width=1\linewidth]{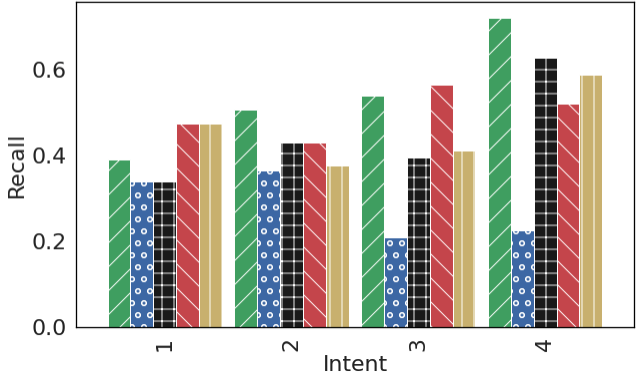}
\caption{Flights Dataset}
\end{subfigure}
\begin{subfigure}{\textwidth}
\includegraphics[width=1\linewidth]{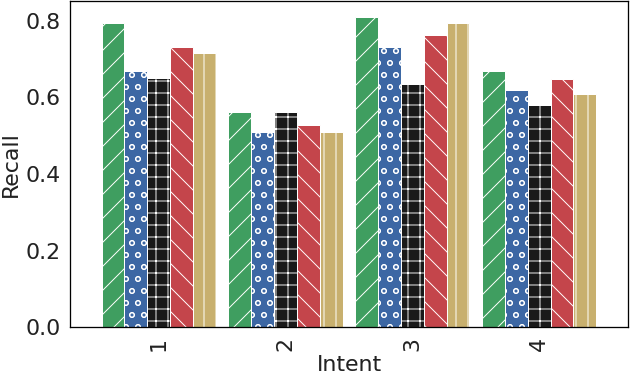}
\caption{Housing Dataset}
\end{subfigure}
\begin{subfigure}{\textwidth}
\includegraphics[width=1\linewidth]{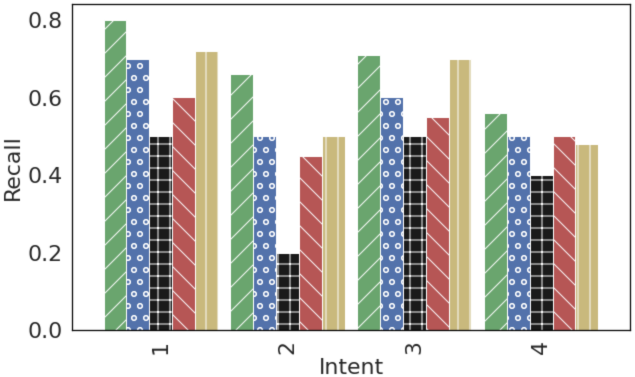}
\caption{Income Dataset}
\end{subfigure}
\caption{Insight Preservation: Recall of the top 5 resulting rows at the end of analysis sequence compared to results from full data, across various intents. Higher is better.}
\label{fig:recall_5}
\end{minipage}%
\hfill
\begin{minipage}[t]{0.32\textwidth}
\centering
\begin{subfigure}{\textwidth}
\includegraphics[width=1\linewidth]{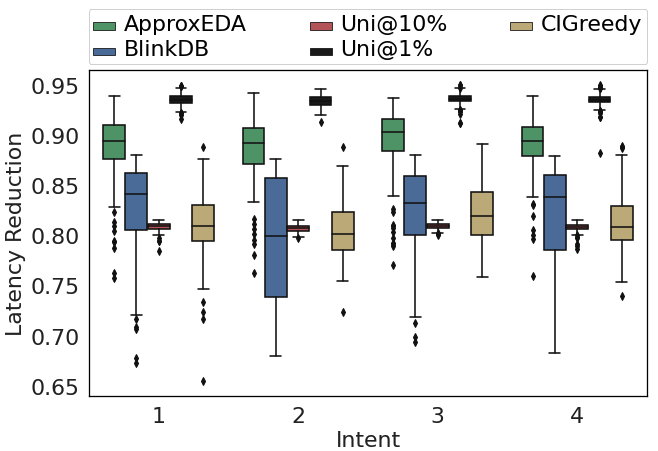}
\caption{Flights Dataset}
\end{subfigure}
\begin{subfigure}{\textwidth}
\includegraphics[width=1\linewidth]{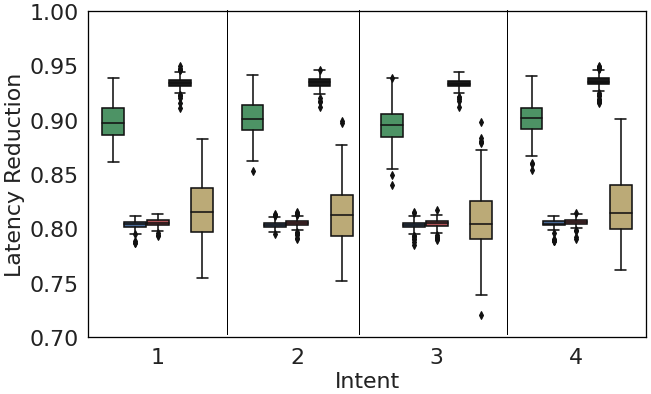}
\caption{Housing Dataset}
\end{subfigure}
\begin{subfigure}{\textwidth}
\includegraphics[width=1\linewidth]{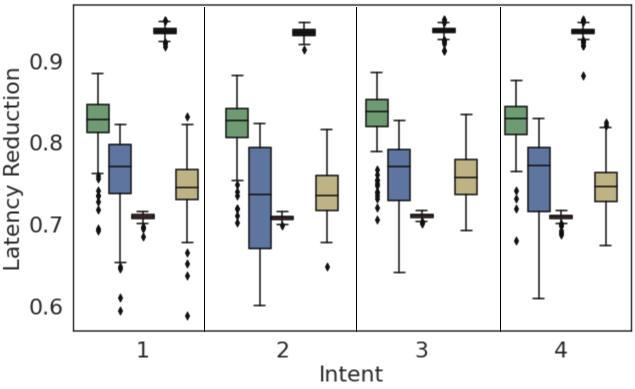}
\caption{Income Dataset}
\end{subfigure}
\caption{The fraction of time saved by queries to run on the sampled dataset when using different models.
     Higher is better.}
     \label{fig:latency_reduction}
\end{minipage}%
\end{minipage}
\hfill
\begin{minipage}[t]{0.29\textwidth}
\centering
\begin{minipage}[t]{1.0\textwidth}
\centering
\scalebox{0.70}{
 \begin{tabular}[t]{|l|l|l|l|}
    \hline
    {Dataset} & {\# rows in full data} & {Unique sequences} \\
    \hline
    {Flights} & {5231403} & {10695}\\
    \hline
    {Housing} & {324196} & {6016} \\
    \hline
    {Income} & {426753} & {4032} \\
    \hline
    \end{tabular}
    }
 \captionof{table}{Dataset and EDA sequences}
 \label{tab:dataset}
\end{minipage}
\begin{minipage}[t]{1.0\textwidth}
\centering
 \includegraphics[width=\textwidth]{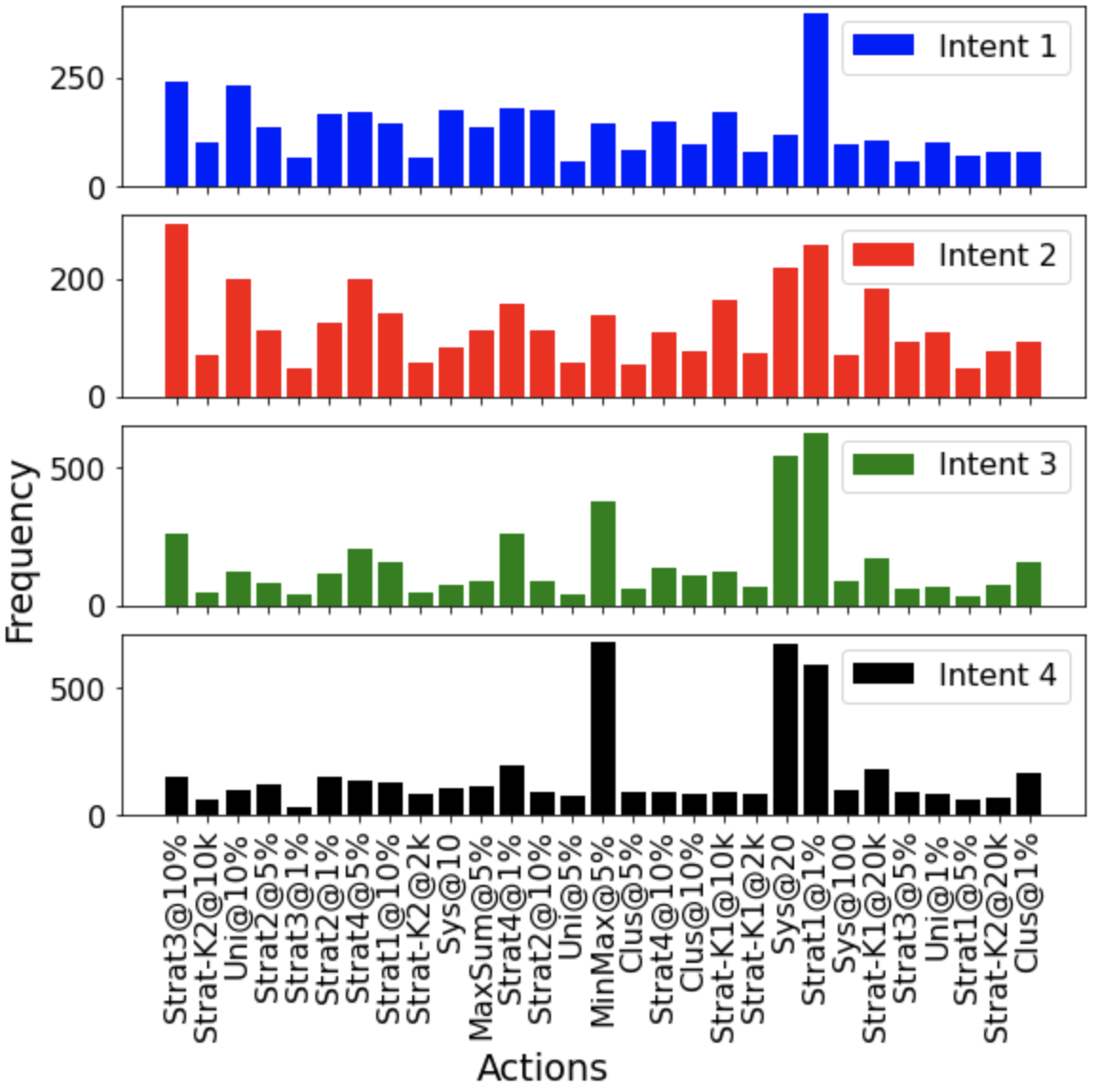}
      \caption{Frequency of different actions taken by \name for different intents.}
      \label{fig:action_usage}
 \end{minipage}
\end{minipage}
\vspace{-0.5em}
\end{figure*}

\noindent\textbf{Datasets.}
%
Table~\ref{tab:dataset} summarizes the popular and public datasets: Flight~\cite{Flights} (also used by \cite{atena2020sigmod}), Housing~\cite{Housing} and Income~\cite{Income}
%
%

\noindent\textbf{Actions}
In total we use 29 actions based on 6 sampling strategies and corresponding parameter combinations as presented in Table~\ref{tab:actions}.
Effective SR denotes the sampling rate defined as total \# rows in the sample / \# rows in the full data. Table~\ref{tab:flights_samples},~\ref{tab:housing_samples},~\ref{tab:income_samples} show the effective sampling rate and the number of rows for different samples for each dataset.

\noindent\textbf{Baselines.}
\begin{itemize}
\item \noindent\textbf{\blink:}
\blink~\cite{agarwal2013blinkdb} attempts to minimize error for each query by selecting the stratified-sample that has the best overlap for QCS (explained before) of the
incoming query to be executed. 
When no overlap with stratified samples, we use 1\% uniform sample. 

\item \noindent\textbf{\greedy:}
We develop another intelligent baseline that first calculates a confidence interval (CI) for aggregate value requested by the incoming query on all the available samples (using a closed form formulation from~\cite{mozafari2015handbook}). 
Then chooses sample with the tightest CI at 95\% (i.e. indicating more confidence in the result).

\item \noindent{Uni@10\% and Uni@1\%:}
We compare our results with two other baselines where the agent always chooses a Uniform 10\% sample and Uniform 1\% sample respectively.
\end{itemize}

\noindent\textbf{Parameters.}
We use $K=4$ intents from BTM as it maximizes overall UCI score. Appendix~\ref{apps:intent_identification} and Figure~\ref{fig:intent_cluster} shows the clusters and UCI scores. We use values of $\beta$ and $\gamma$ as 1 to provide equal weightage to rewards after scaling them.

\subsection {Evaluation Methodology}
\label{sec:evaluation}
\vspace{-0.1em}
%
%
%

\noindent\textbf{EDA Interactions using ATENA.}
%
There are no public datasets available that combines both exploratory insight generation query sequences and the corresponding results on a given data that is also publicly available. 
Moreover, similar to many other interactive scenarios, such as conversational recommender systems \cite{christakopoulou2016towards}, dialog systems \cite{shi2019build,takanobu2020multi} and dialog-based interactive image retrieval \cite{guo2018dialog,yu2019visual}, a major challenge in the evaluation is that we need to have access to user reactions to any possible results returned by the system, which are exhaustive to collect in practice. Therefore, similar to \cite{christakopoulou2016towards,guo2018dialog,yu2019visual,shi2019build}, we adopt a setting where the user reactions are generated by simulators as surrogates for real users.
We overcome this by using an EDA simulator from prior work~\cite{atena2020sigmod}, called ATENA that is trained using real analysis sequences from expert analysts, which uses high-level statistical characteristics 
of the data along with its schema structure 
to generate realistic analysis sequence patterns similar human experts with various implicit intents~\cite{atena2020sigmod}.
For each of the datasets, we train this simulator following the methodology described in \cite{atena2020sigmod}.
Then we use this trained stochastic simulator to generate data exploration sessions which constitute sequence of queries and corresponding results or visualizations. 
For each dataset several such unique sequences can be generated by the simulator using multiple runs and is summarized in Table~\ref{tab:dataset}.
For each dataset, we set aside $1k$ sequences generated by the simulator using full data as held-out set for evaluations.
For \name and the baselines, we also generate $1k$ sequences by letting these methods interact with the simulator where for each query of the EDA, these methods select the optimal samples using their respective algorithms. 
Each of the sequences generated from any of the methods are fed into the BTM model to get the probability distribution of intents over the topics.


\subsection{Evaluation:Intent Divergence (RQ1)}
\label{sec:eval_intent}

We evaluate \name compared to the baselines in preserving the intent distributions (distribution of EDA interaction sequences over topic-models) of the original EDA sequences created without using any samples for approximations. 
In Figure ~\ref{fig:rel_dist_1} we present Euclidean Distance (ED) (normalized w.r.t. \name) between the intent distributions resulting from one of the baselines and from original unapproximated scenario. 
Lower ED is desirable as it indicates better matching of intent distribution. 
We can see that for Flights and Income, \name is best in preserving the distribution. 
For Housing it {Uni@10\%} (uniform sample at 10\%) and \greedy does better. 
This is because both {Uni@10\%} (uniform sample at 10\%) and \greedy ends up choosing large overall sample sizes, and hence less information loss and approximation error.  
However, in the following \S~\ref{sec:eval_latency} and associated latency improvement plot in Figure~\ref{fig:latency_reduction}, we see that such use of large samples makes these methods extremely poor for latency reduction and defeats the purpose of using samples to speed up interactivity. 
Supplement (\textbf{A.1}) provides the exact number of rows corresponding to each samples (actions) created by various sampling strategies.

\begin{figure*}[t!]
\begin{minipage}[t]{0.495\textwidth}
\scalebox{0.67}{
\begin{tabular}{|l|l|l|llll|}
\hline
\multirow{2}{*}{Model} & \multirow{2}{*}{\begin{tabular}[c]{@{}l@{}}Relative\\ ED\end{tabular}} & \multirow{2}{*}{\begin{tabular}[c]{@{}l@{}}Latency\\ Reduction\end{tabular}} & \multicolumn{4}{c|}{Overall Mean Recall}                                                                         \\ \cline{4-7} 
                       &                                                                              &                                                                              & \multicolumn{1}{l|}{Intent 1} & \multicolumn{1}{l|}{Intent 2} & \multicolumn{1}{l|}{Intent 3} & Intent 4 \\ \hline
\name                  & \textbf{1}                                                                            & 0.89                                                                         & \multicolumn{1}{l|}{0.32}     & \multicolumn{1}{l|}{\textbf{0.74}}     & \multicolumn{1}{l|}{0.29}     & 0.63     \\ \hline
\name - $R_{term}$     & 1.08                                                                         & \textbf{0.92}                                                                         & \multicolumn{1}{l|}{\textbf{0.39}}     & \multicolumn{1}{l|}{0.60}     & \multicolumn{1}{l|}{\textbf{0.32}}     & 0.38     \\ \hline
\name - $R_{intent}$   & 1.36                                                                         & 0.91                                                                         & \multicolumn{1}{l|}{0.31}     & \multicolumn{1}{l|}{0.57}     & \multicolumn{1}{l|}{0.30}     & 0.35     \\ \hline
\name - $R_{latency}$  & 1.12                                                                         & 0.83                                                                         & \multicolumn{1}{l|}{0.28}     & \multicolumn{1}{l|}{0.68}     & \multicolumn{1}{l|}{0.29}     & \textbf{0.68}     \\ \hline
\end{tabular}
}
\captionof{table}{Impact of different components of rewards.}
\label{tab:ablation_results}
\end{minipage}%
\hfill
\begin{minipage}[t]{0.49\textwidth}
\scalebox{0.66}{
\begin{tabular}{|l|l|l|llll|}
\hline
\multirow{2}{*}{Action Space}                                             & \multirow{2}{*}{\begin{tabular}[c]{@{}l@{}}Relative\\ ED\end{tabular}} & \multirow{2}{*}{\begin{tabular}[c]{@{}l@{}}Latency\\ Reduction\end{tabular}} & \multicolumn{4}{c|}{Overall Mean Recall}                                                                         \\ \cline{4-7} 
                                                                          &                                                                              &                                                                              & \multicolumn{1}{l|}{Intent 1} & \multicolumn{1}{l|}{Intent 2} & \multicolumn{1}{l|}{Intent 3} & Intent 4 \\ \hline
Only Uniform                                                              & 1.34                                                                         & \textbf{0.90}                                                                         & \multicolumn{1}{l|}{\textbf{0.41}}     & \multicolumn{1}{l|}{0.60}     & \multicolumn{1}{l|}{0.28}     & 0.48     \\ \hline
Uniform + Stratified                                                      & \textbf{0.85}                                                                         & 0.84                                                                         & \multicolumn{1}{l|}{0.34}     & \multicolumn{1}{l|}{0.58}     & \multicolumn{1}{l|}{\textbf{0.30}}     & 0.65     \\ \hline
\begin{tabular}[c]{@{}l@{}}Uniform + Stratified \\ + Cluster\end{tabular} & 1.12                                                                         & \textbf{0.90}                                                                         & \multicolumn{1}{l|}{\textbf{0.41}}     & \multicolumn{1}{l|}{0.60}     & \multicolumn{1}{l|}{\textbf{0.30}}     & \textbf{0.66}     \\ \hline
All Samples                                                               & 1                                                                            & 0.89                                                                         & \multicolumn{1}{l|}{0.32}     & \multicolumn{1}{l|}{\textbf{0.74}}     & \multicolumn{1}{l|}{0.29}     & 0.63     \\ \hline
\end{tabular}
}
\captionof{table}{Ablation study: variations of action space.}
\label{tab:sample_results}
\end{minipage}
\vspace{-0.8em}
\end{figure*}

\vspace{-0.2em}
\subsection{Evaluation: Insight Preservation (RQ2)}
\label{sec:eval_insight}
\vspace{-0.2em}

We evaluate how well the final insight of the EDA is preserved in addition to preserving the intent distribution.
Analysts usually derive insights at the end of each analysis sequence, using the outcome of the final few queries.
Since we do not have explicit labels about whether user actually received satisfactory insights at the end, we calculate a recall value based on results of row obtained from
different sampling based methods w.r.t. results obtained from use of full data.
To calculate this recall, we use top $k=5$ result rows obtained from last 2 queries from each of the analysis sequence -- a thumb rule we understood by talking to experts.

{\footnotesize
\begin{equation}
\label{recall_equation}
    Recall_{Intent=I} = \frac{|\bigcup_{i=1}^{n}M(Q_{i}) \cap \bigcup_{j=1}^{m}M(O_{j})|}{|\bigcup_{j=1}^{m}M(O_{j})|} , {Q_{i},O_{j}} \in I
\end{equation}
}

where $Q_i$ refers to the generated sequence, $O_j$ refers to the sequence generated on full data and $M(Q_i)$ is the union of the top $k$ results of the last two queries of the sequence $Q_i$.

Please note: the intent distribution (\S~\ref{sec:eval_intent}) along with this final insight preservation, as a combination ensures that users' do not diverge from their analysis workflow.  
This is because there can be hypothetical solutions that might directly match the results from last few queries, without replicating the intent of the analysis captured through the entire sequence of analysis. Even though, such a solution  would provide a high recall, but will fail to preserve the purpose of data exploration and associated understanding consumed by analysts~\cite{atena2020sigmod}. 
From Figure~\ref{fig:recall_5} we see that for 2 out of 4 intents for Flights data and all of the intents for Housing data and Income data, \name provides highest recall compared to the baselines. 
As explained in \S~\ref{sec:eval_intent}, the reason either {Uni@10\%} or \greedy sometimes provide higher recall because they end up choosing large sample sizes, which intern lead to much higher latency (Figure~\ref{fig:latency_reduction}). 
We also compute an \textit{overall recall} corresponding to all the results returned by the last 2 queries and illustrate in \ref{app:add_res}.

\vspace{-0.2em}
\subsection{Evaluation:Latency Reduction (RQ3)}
\label{sec:eval_latency}
\vspace{-0.2em}

In Figure~\ref{fig:latency_reduction} we compare the latency improvement by \name compared to the other baselines. 
We compute the overall latency at a sequence level by summing up latency for each individual queries in that sequence. 
We calculate \% of reduction in sequence-level latency compared to the latency when each query execute against full data (y-axis: higher is better).
We show box-plots to show the distribution of values for each sequences, grouped by different intents. 
It can be observed that median sequence-level latency reduction for \name is almost 90\% (that is the analysis sequences take only $1/10^{th}$ of the original time.
Note: even though {Uni@1\%} provides most reduction as it always uses 1\% sample of the data irrespective of the context. 
Recall from Figure~\ref{fig:rel_dist_1} and ~\ref{fig:recall_5}, {Uni@1\%} is pretty bad in terms of preserving the intent distribution or the final insights. The combination of RQ1, RQ2 and RQ3 shows that \name's state can capture different context of sequential analysis and choose optimum trade-off space between statistical effectiveness of different samples vs. corresponding latencies.


\vspace{-0.2em}
\subsection{Qualitative Evaluation}
\vspace{-0.2em}
Due to space, in Appendix~\textbf{\ref{apps:qualitative_eval}} Figure~\ref{fig:real} we show a real EDA session with \name to illustrate in-spite of it selecting frugal samples at various points (e.g. Clus@1\%, Uni@5\% etc.), \name can mostly preserve the query structure, analysis sequence and final outcome/insight. 
In Figure~\ref{fig:action_usage} we show the variation of \name's choice of different actions (selection of samples) for 4 different intents identified for the Flights data across all the held out sequences.
We found that certain samples (e.g. MaxMin@5\%, Sys@20\%, Strat1@1\%) were predominantly chosen for certain intents (e.g. \#4). Intuition is that such samples can preserve representations of smaller groups better, which is necessary for that intent preservation.

\subsection{Ablation Study}
We conduct an ablation study to illustrate the effect of each component on the final metrics in Table~\ref{tab:ablation_results}.
The show Relative-ED, Latency-Reduction and Mean-Overall-Recall as the metric. We progressively remove one reward at a time and see the effect it has on the metrics. Comparing \name and \name - $R_{term}$ , we observe that the latency reduction increases. This is because $R_{latency}$ now has a greater impact on the total reward. However, this is accompanied by a sharp decrease in the Mean Recall value for Intent 2 and Intent 4 as this metric is highly dependant on $R_{term}$. The increase in mean recall for Intent 1 and Intent 3 is nominal.
Removing the $R_{intent}$, we notice that the model now performs worse on every metric except latency. This is because both Relative ED and Mean Recall are highly intent dependant and the lack of intent reward makes it difficult for model to learn a good enough distribution. The improvement in latency can be credited to the higher weight of the latency reward as seen in earlier the earlier case. 
We then verify the effect of removing $R_{latency}$. As expected, we observe that the latency reduction has largely dropped and thus the agent trained without $R_{latency}$ is free to choose larger samples in order to maximise the reward. We notice that, the model now  performs similar to \name and is better in terms of Metric Recall on Intent 4.
Table~\ref{tab:sample_results} shows sensitivity of \name w.r.t. to availability of different groups of sampling algorithms in its action space.

\begin{table}[th]
\tiny

\label{table:ablation_delta}
\begin{tabular}{|l|l|l|llll|}
\hline
\multicolumn{1}{|c|}{\multirow{2}{*}{\begin{tabular}[c]{@{}c@{}}Model \\ Parameter\end{tabular}}} & \multicolumn{1}{c|}{\multirow{2}{*}{Relative ED}} & \multicolumn{1}{c|}{\multirow{2}{*}{\begin{tabular}[c]{@{}c@{}}Latency \\ Reduction\end{tabular}}} & \multicolumn{4}{c|}{\begin{tabular}[c]{@{}c@{}}Mean Overall\\  Recall\end{tabular}}                                           \\ \cline{4-7} 
\multicolumn{1}{|c|}{}                                                                            & \multicolumn{1}{c|}{}                             & \multicolumn{1}{c|}{}                                                                              & \multicolumn{1}{c|}{Intent 1} & \multicolumn{1}{c|}{Intent 2} & \multicolumn{1}{c|}{Intent 3} & \multicolumn{1}{c|}{Intent 4} \\ \hline
$\delta =1$                                                                                         & 1                                                 & 0.89                                                                                               & \multicolumn{1}{l|}{0.32}     & \multicolumn{1}{l|}{0.74}     & \multicolumn{1}{l|}{0.29}     & 0.68                          \\ \hline
$\delta = 0$                                                                                        & 1.2                                               & 0.90                                                                                               & \multicolumn{1}{l|}{0.31}     & \multicolumn{1}{l|}{0.61}     & \multicolumn{1}{l|}{0.3}      & 0.42                          \\ \hline
\end{tabular}
\caption{Ablation study: Variation of $\delta$ in $R_{intent} = R_{dis} + \delta*R_{topic}$ (Ref. \S~\ref{sec:reward}).}
\label{tbl:ablation_delta}
\end{table}

We also do an ablation study on  $\delta$ and show the same in Table~\ref{tbl:ablation_delta}.
We observe that both components of our reward are important. Setting $\delta=0$ increases our relative distance between distributions. Although it reduces the latency, the gain is small. We also see that for various intents, our recall decreases significantly for Intent 2 and Intent 4. This is because setting $\delta=0$ removes the component that matches intent of the original sequence with the generated.

\section{Related Works}
\noindent\textbf{Intent Analysis and Understanding: }
Several papers dealt with the concept of
intents or goals for recommendations in process mining, web mining, education~\cite{jiang2019time, jiang2019goal} and analyze patterns in user behavior from application log/clicks data~\cite{aggarwal2020goal, dev2017identifying} or characterised SQL query logs to understand analysis behaviors~\cite{jain2016sqlshare}. 
Distinct from all these, we identify goal as a combination of queries and the corresponding results in a sequential decision making workflow.

\noindent\textbf{Data Exploration:}
\cite{brachmann2019data} simplified
the manual creation EDA notebooks, cite{chirigati2016data} focused on related dataset recommendation. \cite{zhao2017controlling} attempts to flag false discoveries in EDA automation but does not consider approximations or intent of analysts.
Recent work also focused on automatically generating  data insights~\cite{ma2021metainsight, ding2019quickinsights} or creating or augmenting EDA workflows~\cite{atena2020sigmod, kery2018story, rule2018exploration}. 
Several papers used samples and approximate query processing to speed up data explorations and analytics~\cite{sheoran2022aaai, sheoran2022electra, babcock2003dynamic, chaudhuri2017approximate, ma2021learned, bater2020saqe} but we are the first to look at the sequence and context of such exploration for visual analytics. 

%

\noindent\textbf{Interactive Applications by Reinforcement Learning}
\citet{liu2018dialogue,takanobu2020multi} propose interactive dialog systems based on reinforcement learning.
\citet{guo2018dialog} propose a dialog-based interactive image retrieval system involving natural language feedback on visual attributes of the items.
\citet{tan2019drill} propose reinforcement learning with a drill-down method to interactively retrieve complex scene images.
Reinforcement learning from reformulations is proposed for question answering, by modeling the answering process as multiple agents walking in parallel on the knowledge graph \cite{kaiser2021reinforcement}.  \citet{hua2020few} propose
a meta reinforcement learning method in complex question answering, which quickly adapts to unseen questions. 
Reinforcement learning has also been applied to elicit user preferences in conversational recommender systems \cite{christakopoulou2016towards, zhang2020conversational,lei2020interactive}. Similar to these interactive scenarios \cite{guo2018dialog,yu2019visual,takanobu2020multi}, a major challenge in the evaluation is that we need to have access to user reactions to any possible results returned by the system, which are exhaustive to collect. Therefore, similarly, we adopt a setting where the user reactions are generated by simulators as surrogates for real users as detailed in Section \ref{sec:evaluation}.


\section{Conclusions}

We proposed a deep RL based technique to optimize exploratory data analytics using down-sampled data.  
Our intelligent and contextual sample selection can prevent misleading explorations and insights while still ensuring lower latency of interactions. Evaluations show that our method is superior to baselines when used on three real-world dataset.

\newpage

\vspace{-0.5em}
\bibliography{biblio}

\newpage
\appendix
\clearpage
\section{Supplemental Information}

\begin{algorithm}[th]
        \begin{algorithmic}[1]
        \label{alg:alg1}
        \caption{\name{}'s Training Algorithm}
        \STATE Train the ATENA~\S{5.2} on given dataset $D$
        \STATE Generate sequences $O$ using ATENA
        \STATE Train a $BTM$ model $\phi$ with $K$ topics on $O$
        \STATE Create sample set $A=\{a_1,a_2,...,a_i\}_{i=1}^n$
        \STATE Initialise A2C model
        \FOR{$i = 1,...,episodes$}
            \STATE Initialize the state space $s_t$ as in (\ref{eq:state_space}) with zeroes
            \FOR{$t = 1,...,l$}                    
                    \STATE Generate a query $q_t$ from the simulator
                    \STATE Take action $a_t$ for the given $s_t$
                    \STATE Update $s_t$ with latest query $q_t$, display vectors $v_t$, latency $C_t$ and current intent as $\phi(Q_t)$
            \ENDFOR
            \STATE Calculate Individual rewards using~\S{4.3}.
            \STATE Calculate the total reward $R_t$
            \STATE Calculate $J_R(\pi) = \mathbb{E}_{a_t \sim \pi_\theta}[R_t]$
            \STATE Compute loss functions $\mathbb{L}(\theta)$ for policy network
            \STATE Compute $\mathbb{L}(\omega)$ for value network
            \STATE Update A2C model
        \ENDFOR
        \end{algorithmic}
    \end{algorithm}

\subsection{Formal Problem Formulation}
\label{sec:problem}
%

%
Let $D_0$ be the full data target for EDA.
Let $D'$ = $\{D_i\}_{i=1}^n$ be $n$ subsamples created from $D$ and $|D_i| < |D_0|$ for $i \in [1, n]$ where $|.|$ denotes size (\# of rows). 
Let an $l$ length EDA session $e=\langle \{q_j(D_i), v_j(D_i)\}_{j=1}^l | i \in [1,n]\rangle$ be comprised of an alternating sequence of queries $q_j(.)$ and corresponding results/visual output $v_j(.)$ performed on data $D_i$. 
Note $D$ is the variable $A$ in Section 4.1 and line 4 of Algorithm 1.

Let $I$ be the set of implicit intents embedded in the distribution of all $e$, the EDA sessions.
%
Let $I^{D_0}$ be original distribution of the intents $P(I|D_0)$, i.e. when all EDA was done using full data $D_0$.
Let $I^{D'}$ be the distribution $P(I|D')$, i.e., EDA is approximated using only subsamples.
Our goal is to find optimal subsample $D_i, i \in [1,n]$ for each step of the EDA $\{q_j(.), v_j(.)\}$ such that $Div(I^{D_0}||I^{D'})$ is minimized with $I$ as the support.

We define $Div(I^{D_0}||I^{D'})$ as the \textbf{intent-divergence} between the EDA performed using full data and approximate EDA using only subsampled data. 

\subsection{Discussion: Intent-divergence vs. intent-shift}
\label{app:discussion_intent_shift}

Please note, in this paper we address the problem of \textit{intent-divergence} where the users' have a set of consistent implicit objectives/intents during the EDA. 
While \name{} deliberately introduces a bias towards a set of intents inferred from historical analysis sequences, this is justified as real expert users often follow some standard exploration strategies and look for similar types of insights even on new datasets as observed by multiple prior works~\cite{atena2020sigmod, jain2016sqlshare, milo2018next} and do not perform random explorations.
However, if such implicit objectives suddenly change causing an \textit{intent-shift}, then \name{}'s current design would not be able to detect that. 
In this paper, it is assumed that there is limited \textit{intent-shift} (\emph{i.e.}, change of user's core intent characteristics). 
This assumption is widely adopted in other interactive applications \cite{liu2018dialogue,guo2018dialog,tan2019drill}, although it is interesting to address the intent-shift \cite{xie2021tiage,arnold2019sector}. As the first work on approximate exploratory data analysis in the interactive setting, we leave addressing \textit{intent-shift} as future work.

\begin{figure*}[!t]
\centering
\begin{minipage}[t]{0.57\textwidth}
\vspace{0pt}
\begin{subfigure}{.5\textwidth}
\includegraphics[width=1.0\linewidth]{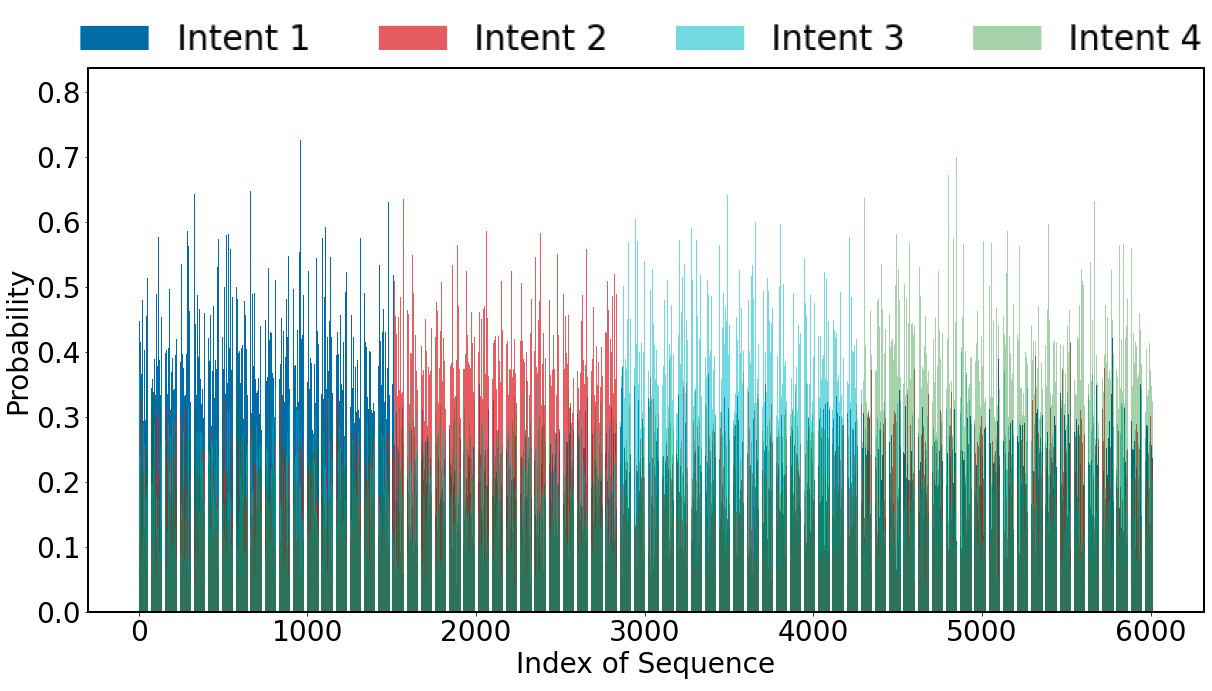}
\caption{EDA on Flights Dataset}
\end{subfigure}
\begin{subfigure}{.5\textwidth}
\includegraphics[width=1.0\linewidth]{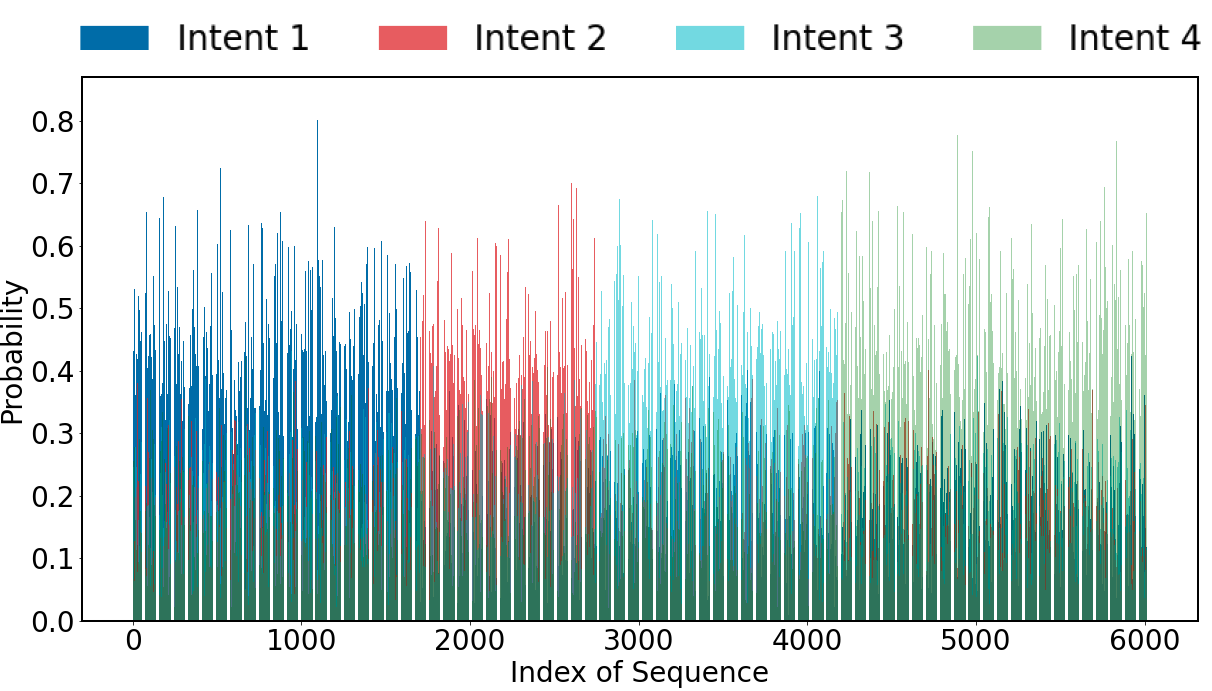}
\caption{EDA on Housing Dataset}
\end{subfigure}
     \caption{The probability distribution of query sequences for each of the identified $k=4$ intents. The sequences are grouped together intent-wise. This graph illustrates the distinguishable definitions of intents through their probability distributions. Since trend on Income dataset is similar, we omit the details.}
     \label{fig:intent_cluster}
\end{minipage}
\hspace{0.05\textwidth}
\begin{minipage}[t]{0.37\textwidth}
\vspace{0pt}
\begin{subfigure}{1.0\textwidth}
\includegraphics[width=1.0\linewidth]{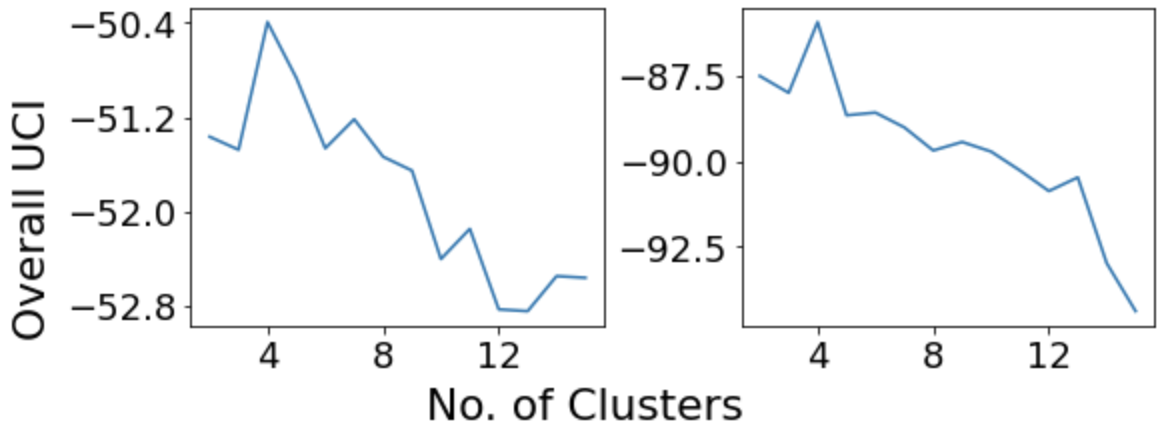}
\end{subfigure}
\caption{UCI measure for different number of clusters (Left: Flights, Right: Housing data). Overall UCI is the highest when $K=4$ for both. Since trend on Income dataset is similar, we omit the details.}
     \label{fig:intent_clusters}
\end{minipage}
\vspace{-0.5em}
\end{figure*}

\subsection{Discussions: Design and Deployment}
\label{app:discussion}

\noindent\textbf{RL vs. Greedy Solutions.}

Simple greedy solutions would not work because because of the following reasons:
(1) It is impossible to calculate an approximation error when a query is run against each of the samples, unless we also run it against the original data at run-time (which defeats the purpose of using samples), 
(2) Even if we calculate an estimate of the approximation error, choosing the sampling strategy that minimizes error for each query may not be optimal because a less-accurate strategy might be of significantly lesser size, but still contain enough information that would preserve the high-level signal in the results (e.g. a bar being substantially higher than the rest) and would not lead to a wrong decisions by the analyst while choosing subsequent queries.

\noindent\textbf{Deployment Aspects.}

Significant history of EDA analysis sequence gets generated for data analysis usecases related to e-commerce, web/mobile analytics or digital marketing setting, where several analysts repeatedly explore the data to constantly understand different aspects of customer behavior. 
%
In production scenario, different samples  are periodically regenerated in an offline manner as new batch of data of significant size is ingested. 
However, according to our understanding from experts, the EDA analysis patterns and intents for a particular dataset remains same over time (also refer to \S~\ref{app:discussion_intent_shift}).
Since, the overall characteristics of intents of the analysts do not change, \name{}'s learning about the effectiveness of different samples under different contexts does not get significantly affected if the overall statistical characteristics of the freshly ingested data is consistent with the old one.
However, if the statistical characteristics change, then \name{} is retrained with new samples in the background and after convergence the old model is replaced.

\subsection{More Ablation Study Results}

\begin{table}[th]
\tiny

\label{table:ablation_delta}
\begin{tabular}{|l|l|l|llll|}
\hline
\multicolumn{1}{|c|}{\multirow{2}{*}{\begin{tabular}[c]{@{}c@{}}Model \\ Parameter\end{tabular}}} & \multicolumn{1}{c|}{\multirow{2}{*}{Relative ED}} & \multicolumn{1}{c|}{\multirow{2}{*}{\begin{tabular}[c]{@{}c@{}}Latency \\ Reduction\end{tabular}}} & \multicolumn{4}{c|}{\begin{tabular}[c]{@{}c@{}}Mean Overall\\  Recall\end{tabular}}                                           \\ \cline{4-7} 
\multicolumn{1}{|c|}{}                                                                            & \multicolumn{1}{c|}{}                             & \multicolumn{1}{c|}{}                                                                              & \multicolumn{1}{c|}{Intent 1} & \multicolumn{1}{c|}{Intent 2} & \multicolumn{1}{c|}{Intent 3} & \multicolumn{1}{c|}{Intent 4} \\ \hline
$\delta =1$                                                                                         & 1                                                 & 0.89                                                                                               & \multicolumn{1}{l|}{0.32}     & \multicolumn{1}{l|}{0.74}     & \multicolumn{1}{l|}{0.29}     & 0.68                          \\ \hline
$\delta = 0$                                                                                        & 1.2                                               & 0.90                                                                                               & \multicolumn{1}{l|}{0.31}     & \multicolumn{1}{l|}{0.61}     & \multicolumn{1}{l|}{0.3}      & 0.42                          \\ \hline
\end{tabular}
\caption{Ablation study: Variation of $\delta$ in $R_{intent} = R_{dis} + \delta*R_{topic}$ (Ref. \S~\ref{sec:reward}).}
\label{tbl:ablation_delta}
\end{table}

Recall in \S~\ref{sec:reward}, we use full intent reward as: $R_{intent} = R_{dis} + \delta*R_{topic}$.
We do an ablation study on this $\delta$ and show in Table~\ref{tbl:ablation_delta}.
We observe that both components of our reward are important. Setting $\delta=0$ increases our relative distance between distributions. Although it reduces the latency, the gain is small. We also see that for various intents, our recall decreases significantly for Intent 2 and Intent 4. This is because setting $\delta=0$ removes the component that matches intent of the original sequence with the generated.

\begin{table}[]
\scalebox{0.8}{
\begin{tabular}{|l|l|}
\hline
No. of Intents & \begin{tabular}[c]{@{}l@{}}Relative  ED\end{tabular} \\ \hline
4              & 1                                                     \\ \hline
6              & 2.98                                                  \\ \hline
8              & 5.84                                                  \\ \hline
10             & 7.41                                                  \\ \hline
\end{tabular}
}
\caption{Ablation study: Variation of K or \# of intents (Ref ~\S{\ref{sec:intent_identification}})}
\label{tbl:ablation_K_intent}
\end{table}

In Table~\ref{tbl:ablation_K_intent} we show ablation study for Flights dataset on number of intents used ($K$) (Ref. ~\S{\ref{sec:intent_identification}}) and how Euclidean Distance (ED) from ideal changes (increase in ED is worse) relative to the distance for number of intents $K=4.$


\subsection{Intent Identification Results} 
\label{apps:intent_identification}

Figure~\ref{fig:intent_cluster} shows the probability distribution of intents for each of the query sequence and the 4 clusters identified can be clearly observed for Flights and Housing datasets. 
Since trend on Income dataset is similar, we omit the details.
To choose the best value for the number of intents, $K$, we compute the overall UCI score for all $K \in [2,15]$ and choose the best K. As can be seen maximum UCI score is observable at K=4.

\subsection{Model and Implementation Details}
\label{apps:model_implementation}
Our approach is based on A2C, an actor critic model for the training of our RL Agent. An actor critic model consists of two parts, an actor (policy network) which takes the state of the agent and outputs the best possible action, whereas the critic (value network) evaluates the action by computing the value function. 


To train our model, we generate 10k sequences using the simulator~\cite{atena2020sigmod} and we get 5995 unique sequences for the flights dataset and 6016 sequences for the housing dataset. We use $K=4$ for our experiments. To train the RL agent, for both policy network and value network, we use neural networks consisting of 2 fully-connected layers, with a dimension of 64 each. The activation function used is $\tanh$. The policy model used is MlpPolicy and we set the number of environments to 64. For the A2C model, we set the value function coefficient, $vf_{coef} = 0.25$ and the entropy coefficient, $ent_{coef} = 0.01$. The learning rate used for the model is 0.0007. More details about the algorithm are explained in \S{\ref{app:training}}
 
For the reward calculation, we scale all the rewards between $[-0.5, 0.5]$. For training \name{}, we used $\delta=1, \zeta=1$. To give equal weight to all the reward components, we set the value of $\beta=0.5, \gamma=0.5$. \ref{tab:model-details} shows some details for the trained model.

\begin{figure}[h]

    \includegraphics[width=1\linewidth]{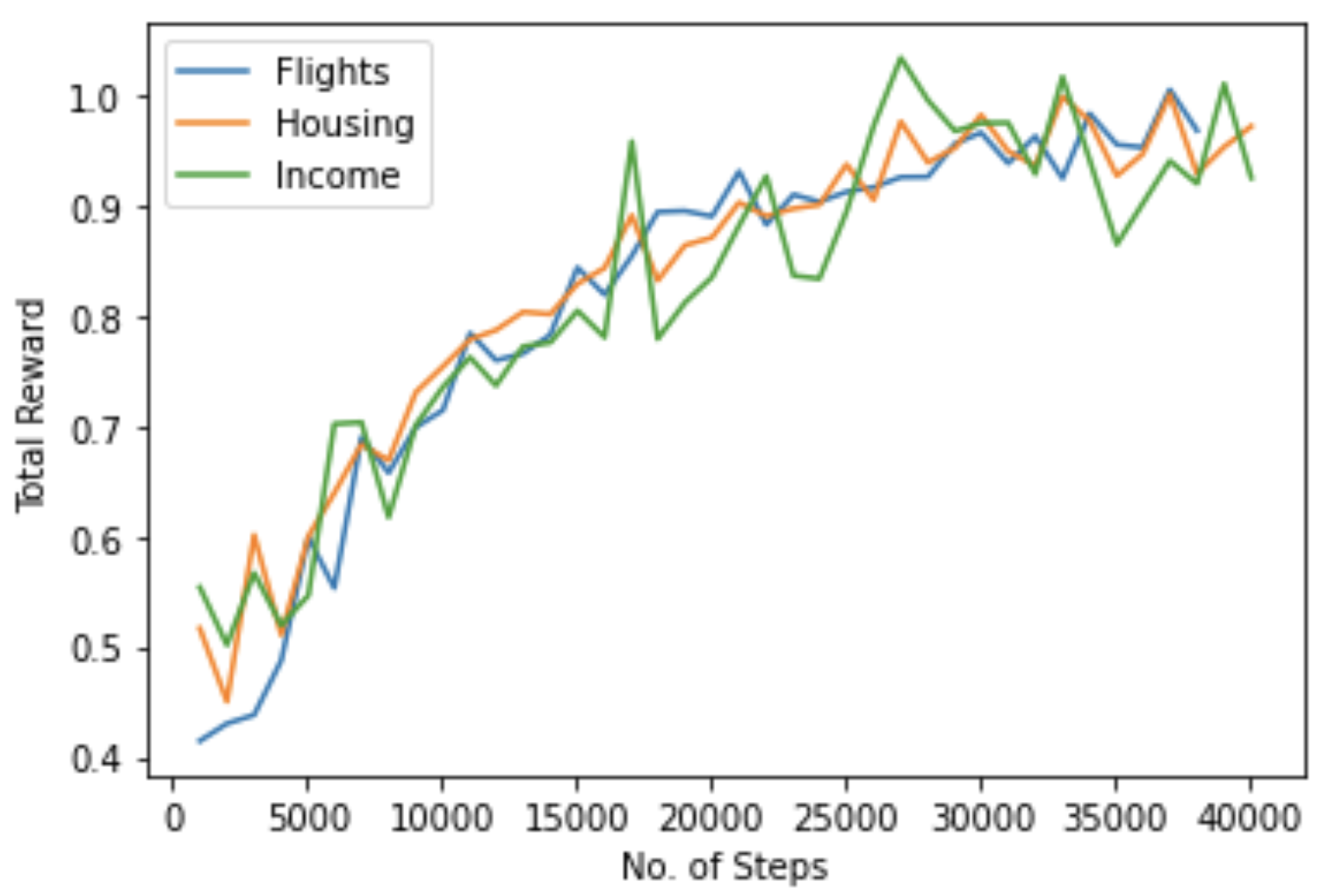}
    \caption{RL training convergence curves for different datasets}
    \label{fig:my_conv}

\end{figure}

Tables \ref{tab:flights_samples}, \ref{tab:housing_samples} and \ref{tab:income_samples} show the actual number of rows sampled from each dataset and presents the Effective Sampling Rate (ESR). We create 29 samples for the \textit{Flights dataset} as shown and create 15 samples for each of \textit{Housing dataset} and \textit{Income dataset}.
In Figure~\ref{fig:my_conv} we show the training convergence curves. For evaluation, we choose the model with the best possible reward. The no. of steps for selection of the model is given in Table 7.

\begin{table}[H]
\vspace{0pt}
\scalebox{1}{
\begin{tabular}{|l|l|l|l|}
\hline
Dataset & \begin{tabular}[c]{@{}l@{}}No. of \\ steps\end{tabular} & Time(hr) & Model Size(KiB) \\ \hline
Housing & 36k                                                       & 20.3     & 284       \\ \hline
Flights & 36k                                                       & 15.8     & 363       \\ \hline
Income & 33k                                                       & 16.7     & 312       \\ \hline
\end{tabular}
}
\caption{Training times and model sizes}
\label{tab:model-details}
\end{table}
 
\subsection{Training Algorithm}
\label{app:training}
Algorithm 1 explains the steps we use for training our RL Agent. It takes in the dataset $D$, processes it by creating samples as shown in line 4 and training ATENA simulator on the same as shown in line 1. Then on the generated sequences, a BTM model is trained in line 3. Once this is done, we then start the training of our agent as demonstrated in lines 5-18 of Algorithm 1. We follow episodic training where we calculate rewards at the end of each episode as shown in line 13. Once we calculate the total reward, we compute the loss functions and update both actor (policy) and critic (value) networks as explained in \S{\ref{apps:model_implementation}}. 

\subsection{Qualitative Evaluation}
\label{apps:qualitative_eval}
%
\begin{figure}[th!]
\begin{minipage}[t]{1.0\columnwidth}
\vspace{0pt}
\includegraphics[width=1.0\columnwidth]{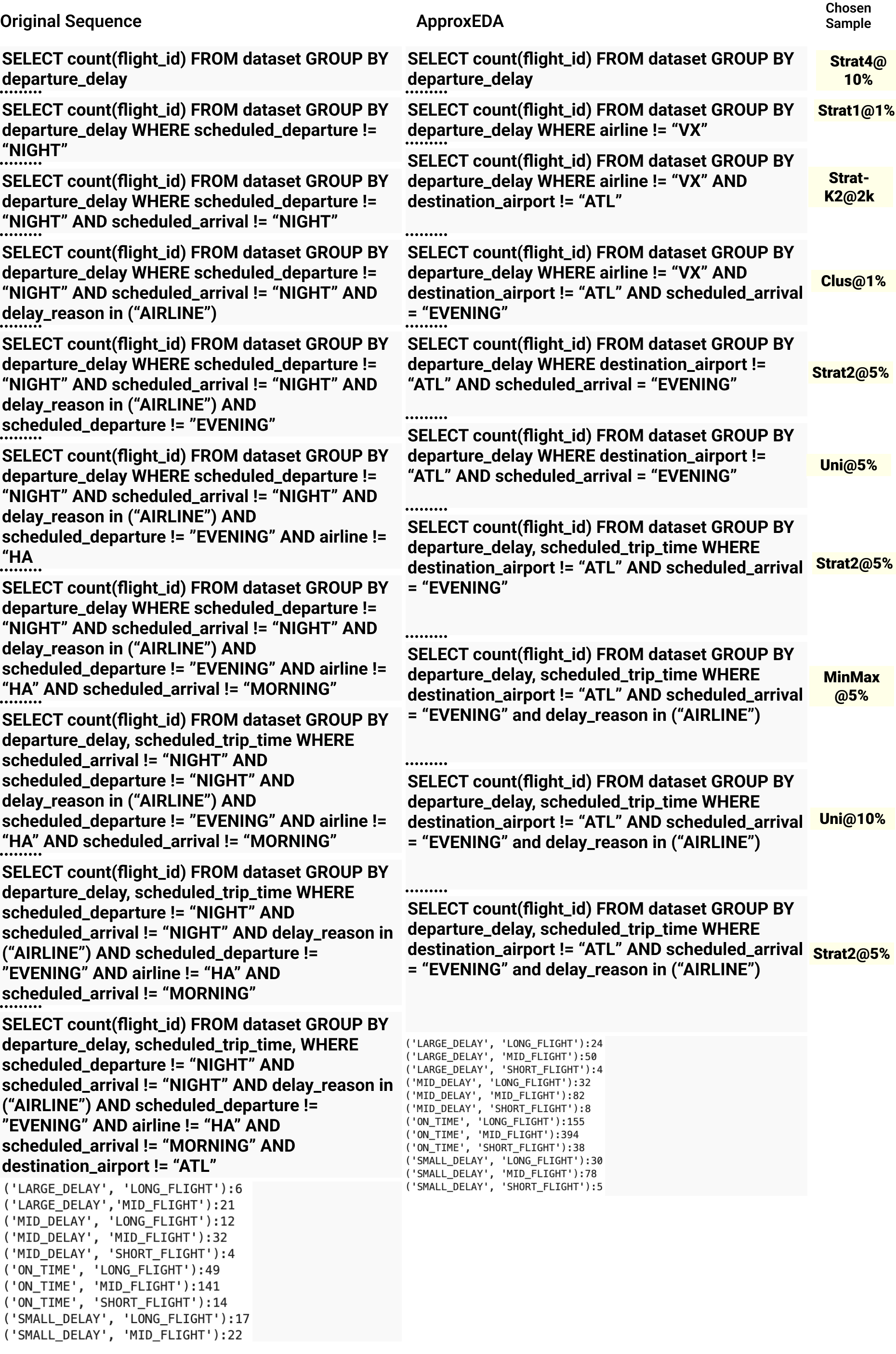}
     \caption{
     An example EDA sequence from the logged results,
     showing how \name{} can efficiently preserve the query structure, analysis sequence and final outcome/insight by selecting various different samples (annotations on the right) under the hood to maximize its objective.}
     \label{fig:real}
\end{minipage}
\end{figure}

Figure~\ref{fig:real} we show a real EDA session with \name{}, where it 
selected various different samples (annotations on the right) to maximize its objective. We can see in-spite of it selecting fery frugal samples at various points (e.g. Clus@1\%, Uni@5\% etc.), \name{} can mostly preserve the query structure, analysis sequence and final outcome/insight. 
In Figure~\ref{fig:action_usage} we show the variation of \name{}'s choice of different actions (selection of samples) for 4 different intents identified for the Flights data across all the held out sequences.
We found that certain samples (e.g. MaxMin@5\%, Sys@20\%, Strat1@1\%) were predominantly chosen for certain intents (e.g. \#4). Intuition is that such samples can preserve representations of smaller groups better, which is necessary for that intent preservation.


\subsection{Additional Results}
\label{app:add_res}
Similar to \ref{sec:eval_insight}, we also compute the recall for all the results instead of just 5. We show the results in \ref{fig:recall}. Even in the case of the whole recall \name{} seems to do better than the baselines.

\begin{figure}[!t]
\centering
\begin{minipage}[t]{0.495\columnwidth}
\vspace{0pt}
\scalebox{0.65}{
\begin{tabular}{|l|l|l|}
\hline
\begin{tabular}[c]{@{}l@{}}Sampling \\ Strategy\end{tabular} & \begin{tabular}[c]{@{}l@{}}No. of \\ Rows\end{tabular} & \begin{tabular}[c]{@{}l@{}}Effective\\  SR\end{tabular} \\ \hline
Uni@1\%                                                      & 5232                                                   & 0.01                                                    \\ \hline
Uni@5\%                                                      & 26155                                                  & 0.05                                                    \\ \hline
Uni@10\%                                                     & 52312                                                  & 0.1                                                     \\ \hline
Sys@100                                                      & 5231                                                   & 0.01                                                    \\ \hline
Sys@20                                                       & 26154                                                  & 0.05                                                    \\ \hline
Sys@10                                                       & 52311                                                  & 0.1                                                     \\ \hline
Clus@1\%                                                     & 5230                                                   & 0.01                                                    \\ \hline
Clus@5\%                                                     & 26154                                                  & 0.05                                                    \\ \hline
Clus@10\%                                                    & 52306                                                  & 0.1                                                     \\ \hline
MinMax@5\%                                                   & 27202                                                  & 0.052                                                   \\ \hline
MaxSum@5\%                                                   & 25109                                                  & 0.048                                                   \\ \hline
KStrat1@2k                                                   & 14124                                                  & 0.027                                                   \\ \hline
KStrat1@10k                                                  & 28771                                                  & 0.055                                                   \\ \hline
KStrat1@20k                                                  & 57542                                                  & 0.11                                                    \\ \hline
KStrat2@2k                                                   & 9416                                                   & 0.018                                                   \\ \hline
KStrat2@10k                                                  & 25125                                                  & 0.048                                                   \\ \hline
KStrat2@20k                                                  & 47080                                                  & 0.09                                                    \\ \hline
Strat1@1\%                                                   & 5220                                                   & 0.01                                                    \\ \hline
Strat1@5\%                                                   & 26146                                                  & 0.05                                                    \\ \hline
Strat1@10\%                                                  & 52375                                                  & 0.1                                                     \\ \hline
Strat2@1\%                                                   & 5232                                                   & 0.01                                                    \\ \hline
Strat2@5\%                                                   & 26148                                                  & 0.05                                                    \\ \hline
Strat2@10\%                                                  & 52320                                                  & 0.1                                                     \\ \hline
Strat3@1\%                                                   & 5240                                                   & 0.01                                                    \\ \hline
Strat3@5\%                                                   & 26150                                                  & 0.05                                                    \\ \hline
Strat3@10\%                                                  & 52315                                                  & 0.1                                                     \\ \hline
Strat4@1\%                                                   & 5235                                                   & 0.01                                                    \\ \hline
Strat4@5\%                                                   & 26149                                                  & 0.05                                                    \\ \hline
Strat4@10\%                                                  & 52321                                                  & 0.1                                                     \\ \hline
\end{tabular}
}
\captionof{table}{Details of Samples generated on Flights Dataset}
\label{tab:flights_samples}
\end{minipage}%
\hfill
\begin{minipage}[t]{0.495\columnwidth}
\vspace{0pt}
\begin{minipage}[t]{1.0\columnwidth}
\begin{minipage}[t]{1.0\columnwidth}
\vspace{0pt}
\scalebox{0.65}{
\begin{tabular}{|l|l|l|}
\hline
\begin{tabular}[c]{@{}l@{}}Sampling \\ Strategy\end{tabular} & \begin{tabular}[c]{@{}l@{}}No. of \\ Rows\end{tabular} & \begin{tabular}[c]{@{}l@{}}Effective\\  SR\end{tabular} \\ \hline
Uni@1\%                                                      & 3242                                                   & 0.01                                                    \\ \hline
Uni@5\%                                                      & 16210                                                  & 0.05                                                    \\ \hline
Uni@10\%                                                     & 32419                                                  & 0.1                                                     \\ \hline
Sys@100                                                      & 3241                                                   & 0.01                                                    \\ \hline
Sys@20                                                       & 16209                                                  & 0.05                                                    \\ \hline
Sys@10                                                       & 32419                                                  & 0.1                                                     \\ \hline
Strat1@1\%                                                   & 3246                                                   & 0.01                                                    \\ \hline
Strat1@5\%                                                   & 16225                                                  & 0.05                                                    \\ \hline
Strat1@10\%                                                  & 32425                                                  & 0.1                                                     \\ \hline
Strat2@1\%                                                   & 3238                                                   & 0.01                                                    \\ \hline
Strat2@5\%                                                   & 16215                                                  & 0.05                                                    \\ \hline
Strat2@10\%                                                  & 32426                                                  & 0.1                                                     \\ \hline
Strat3@1\%                                                   & 3240                                                   & 0.01                                                    \\ \hline
Strat3@5\%                                                   & 16208                                                  & 0.05                                                    \\ \hline
Strat3@10\%                                                  & 32420                                                  & 0.1                                                     \\ \hline
Strat4@1\%                                                   & 3248                                                   & 0.01                                                    \\ \hline
Strat4@5\%                                                   & 16206                                                  & 0.05                                                    \\ \hline
Strat4@10\%                                                  & 32425                                                  & 0.1                                                     \\ \hline
\end{tabular}
}

\captionof{table}{Details of Samples generated on Housing Dataset}
\label{tab:housing_samples}
\end{minipage}
\begin{minipage}[t]{1.0\columnwidth}
\scalebox{0.65}{
\begin{tabular}{|l|l|l|}
\hline
\begin{tabular}[c]{@{}l@{}}Sampling \\ Strategy\end{tabular} & \begin{tabular}[c]{@{}l@{}}No. of \\ Rows\end{tabular} & \begin{tabular}[c]{@{}l@{}}Effective\\  SR\end{tabular} \\ \hline
Uni@1\%                                                      & 4266                                                   & 0.01                                                    \\ \hline
Uni@5\%                                                      & 21334                                                  & 0.05                                                    \\ \hline
Uni@10\%                                                     & 42670                                                 & 0.1                                                     \\ \hline
Sys@100                                                      & 4266                                                  & 0.01                                                    \\ \hline
Sys@20                                                       & 21334                                                & 0.05                                                    \\ \hline
Sys@10                                                       & 42669                                                  & 0.1                                                     \\ \hline
Strat1@1\%                                                   & 4267                                                   & 0.01                                                    \\ \hline
Strat1@5\%                                                   & 21336                                                  & 0.05                                                    \\ \hline
Strat1@10\%                                                  & 42672                                                  & 0.1                                                     \\ \hline
Strat2@1\%                                                   & 4235                                                   & 0.01                                                    \\ \hline
Strat2@5\%                                                   & 21330                                                  & 0.05                                                    \\ \hline
Strat2@10\%                                                  & 42664                                                  & 0.1                                                     \\ \hline
Strat3@1\%                                                   & 4235                                                   & 0.01                                                    \\ \hline
Strat3@5\%                                                   & 21340                                                  & 0.05                                                    \\ \hline
Strat3@10\%                                                  & 42667                                                  & 0.1                                                     \\ \hline
Strat4@1\%                                                   & 4237                                                   & 0.01                                                    \\ \hline
Strat4@5\%                                                   & 21367                                                  & 0.05                                                    \\ \hline
Strat4@10\%                                                  & 42673                                                  & 0.1                                                     \\ \hline
\end{tabular}
}

\captionof{table}{Details of Samples generated on Income Dataset}
\label{tab:income_samples}
\end{minipage}
\end{minipage}
\end{minipage}
\end{figure}


\subsection{Sampling Strategies}
\label{app:sampling_details}

A sample dataset $T_s$ is a subset of rows from an original dataset $T$. Following different sampling strategies select these subsets differently with different set of parameters that control the size and statistics of $T_s$.

\noindent \textbf{Uniform Random Sampling:}
In uniform sampling, the rows of $T$ are
sampled independently (i.e, a Bernoulli process) with a sampling probability equal to $\tau \in [0,1] $. $\tau$ is called the sampling frequency and different values lead to different sizes for $T_s$.

\noindent \textbf{Systematic Sampling:}
In a systematic sample~\cite{lohr2019sampling}, a starting row of data is chosen using a random number. That row, and every $k$-th row thereafter, is chosen to be in $T_s$.

\noindent \textbf{Stratified Sampling:}
Let for a column $C$ in data $T$, there are $G$ unique categories (i.e. strata). 
Stratified sample ensures that \textit{enough} rows are selected in $T_s$ for each unique values
in $G$. 
%


\noindent\textbf{Proportional stratified sample:} 
    In this case uniform samples $T_{i_{s}}$ are picked for each strata $T_{i}$ with a sampling probability of $\tau$.
    
    
\noindent\textbf{At most $K$ stratified sample:}
    In this case, for each strata, at most $K$ values are randomly selected to be included in $T_s$~\cite{agarwal2013blinkdb}.
    

\noindent \textbf{Cluster Sampling:}
The data is clustered into $k$  clusters,
samples are picked from each cluster with a sampling probability of $\tau$.


\noindent \textbf{Diversity sampling:} 
This targets that the sample picked out has different types of rows inside~\cite{diversity2017}. 
%
%
Diversity is defined by a distance function $d$, including: (i) \textbf{MaxMin Diversity:} 
    This maximises the minimum pairwise diversity between elements of $T_s$. 
    %
and (ii)  \textbf{MaxSum Diversity:}
    This maximises the average pairwise diversity between elements of $T_s$.

Tables \ref{tab:flights_samples}, \ref{tab:housing_samples} and \ref{tab:income_samples} show the sampling strategies and corresponding parameters used, actual size of the samples ($|T_s|)$ and presents the Effective Sampling Rate (ESR) for each of the three datasets we used to evaluate. Lower the sample size, lower would be the query latency but at the cost of different magnitude of approximation errors depending on the actual sampling strategy. 

%
\end{document}